%% file: PSP-DIT.tex
\documentclass{article}

\usepackage[main, final]{neurips_2026}

\usepackage[utf8]{inputenc} 
\usepackage[T1]{fontenc}    
\usepackage[hypertexnames=false]{hyperref}       
\usepackage{xurl}           
\usepackage{url}            
\usepackage{booktabs}       
\usepackage{amsfonts}       
\usepackage{nicefrac}       
\usepackage{microtype}      
\usepackage{xcolor}         
\usepackage{array}
\usepackage{makecell}
\newcolumntype{L}[1]{>{\raggedright\arraybackslash}p{#1}}
\newcolumntype{C}[1]{>{\centering\arraybackslash}m{#1}}
\usepackage{graphicx}
\usepackage{amsmath}
\usepackage{enumitem}
\usepackage{algorithm}
\usepackage{algpseudocode}
\usepackage{float}
\hypersetup{
    colorlinks=true,
    linkcolor=[rgb]{0.18,0.35,0.60},
    citecolor=[rgb]{0.18,0.35,0.60},
    urlcolor=[rgb]{0.18,0.35,0.60},
    pdfborder={0 0 0}
}
\title{Panoptic Scene Program Diffusion Transformer}

\author{%
  Chika Maduabuchi \\
  William \& Mary \\
}

\begin{document}

\maketitle

\input{sections/abstract}

\begin{figure}[!t]
    \centering
    \includegraphics[width=\linewidth]{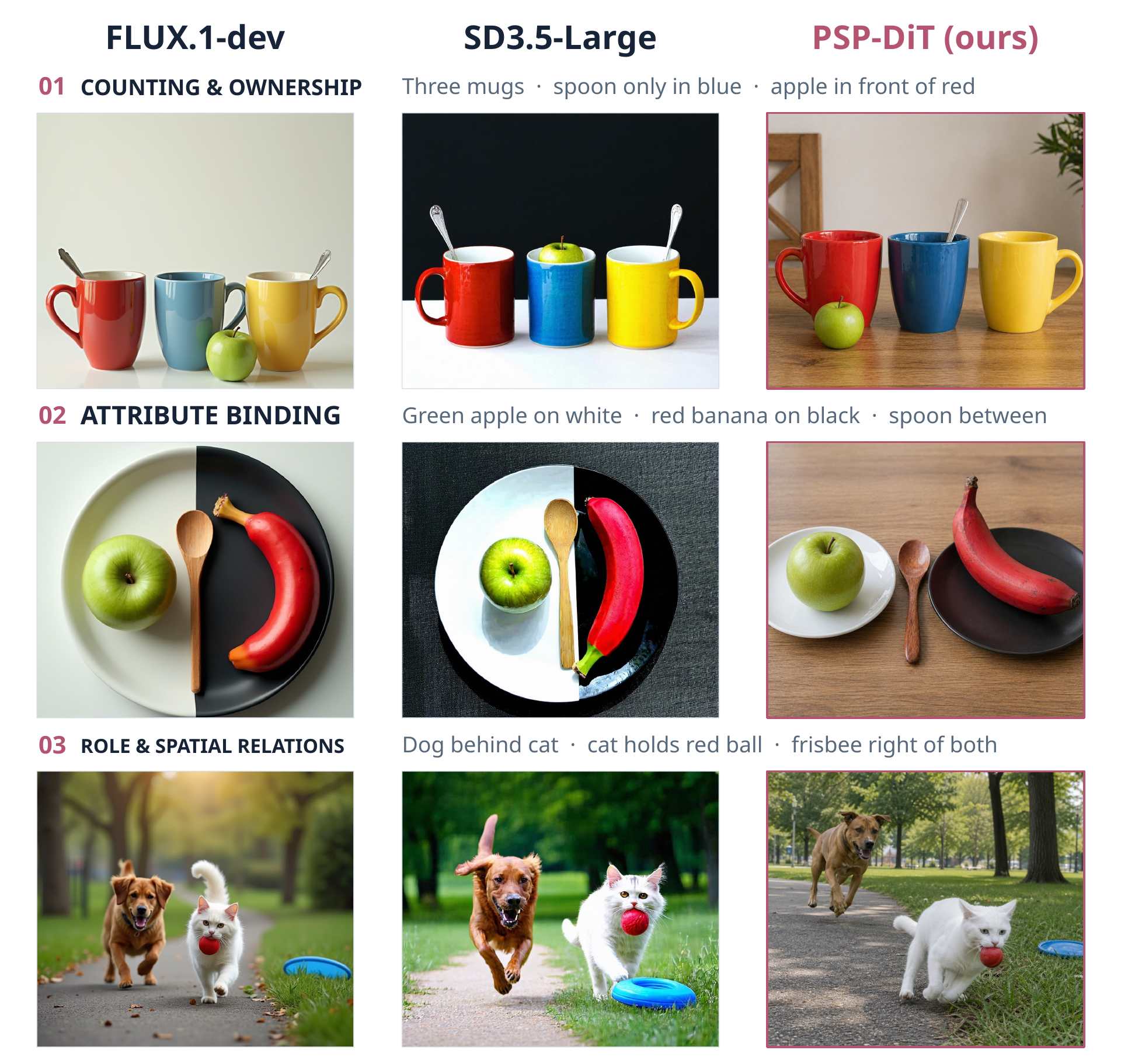}
    \caption{Qualitative comparisons of FLUX.1-dev,
    SD3.5-Large, and PSP-DiT on three shared
    compositional prompts. The rows illustrate counting
    and ownership, attribute binding, and role-sensitive
    spatial relations. Additional annotated comparisons
    are discussed in Section~4.9.}
    \label{fig:teaser}
\end{figure}

\input{sections/introduction}
\input{sections/related}
\input{sections/methods}
\input{sections/results}
\input{sections/conclusions}

\begin{ack}
The author acknowledges computational support from the NSF-supported Delta and DeltaAI systems.

\noindent\textbf{Competing Interests.} The author declares no competing interests relevant to this work.
\end{ack}

\clearpage

\bibliographystyle{plainnat}
\bibliography{reference}

\newpage

\input{sections/appendix}

\clearpage
\input{checklist.tex}

\end{document}

%% file: sections/abstract.tex
\begin{abstract}
Modern text-to-image models produce high-fidelity images but still struggle with compositional prompts that require instance identity, attribute ownership, counting, spatial ordering, and role-sensitive relations. We introduce \emph{Panoptic Scene Program Diffusion Transformer} (PSP-DiT), a diffusion-transformer architecture that treats a panoptic scene program as a first-class latent variable rather than an external control signal or post-hoc parse. PSP-DiT jointly denoises image latents and scene-program latents through coupled transformer streams, while panoptic grounding and cycle-consistency objectives tie object instances, attributes, relations, and counts to visual support in the generated image. Under matched training and inference settings, PSP-DiT improves over a strong flat-text baseline across GenEval 2, SANEval-Simple, PSG-Score, and DetailMaster, with the largest gains on counting, attribute binding, role-sensitive relations, and long structured prompts. The method preserves image quality, adds modest inference overhead, and remains robust to imperfect scene programs.
\end{abstract}

%% file: sections/introduction.tex
\section{Introduction}
\label{sec:introduction}

Text-to-image generation has advanced at extraordinary speed, evolving from latent diffusion and large language-conditioned cascades to diffusion transformers and rectified-flow systems that now deliver strong visual fidelity, prompt adherence, and typography at unprecedented scale~\citep{rombach2022high,saharia2022photorealistic,betker2023dalle3,peebles2023scalable,esser2024srf,xie2024sana,tong2026scaleRAE}. Yet the central scientific bottleneck has shifted rather than disappeared: the hardest failures are no longer low-level realism, but compositional faithfulness under multi-entity, detail-heavy prompts. Across benchmarks spanning object co-occurrence, counting, attribute binding, spatial relations, long prompts, spatial intelligence, and reasoning, state-of-the-art systems still degrade precisely when they must keep track of instance identity, ownership, ordering, and interaction structure~\citep{ghosh2023geneval,huang2023t2icompbench,kamath2025geneval2,pramanik2026saneval,jiao2025detailmaster,wang2026everything,chen2025r2ibench,zhang2025worldgenbench,gao2024generateanyscene}. This pattern suggests that the remaining barrier is not only stronger text encoders or more data, but the absence of an explicit scene state inside the generative process itself.

A broad line of work has tried to close this gap through controllable or structured generation. Training-time and inference-time guidance methods such as GLIGEN, ControlNet, Attend-and-Excite, and BoxDiff improve controllability by injecting boxes, masks, or attention constraints into pretrained diffusion models~\citep{li2023gligen,zhang2023controlnet,kirstain2023attendexcite,xie2023boxdiff}. Scene-graph and structure-centric approaches further demonstrate the value of relational representations for generation, evaluation, and data synthesis~\citep{xu2024diffusesg,wang2024scenegraph,chen2024scenebench,gao2024generateanyscene,jayasekara2026allinone}. In parallel, panoptic scene graph research has shown that instance-resolved, region-grounded relational representations are substantially richer than box-level graphs for visual understanding~\citep{yang2022psg,zhao2023caption2psg,zhou2024openpsg}. However, these directions remain fundamentally separated: control methods treat structure as an external condition, scene-graph methods typically factor graph generation from image generation or translate graphs into layouts and semantics used only as guidance, and panoptic scene graphs are studied primarily for recognition rather than as first-class latent variables in modern text-to-image generators. We target this missing middle. Our method, \emph{Panoptic Scene Program Diffusion Transformer} (PSP-DiT), makes an instance-indexed panoptic scene program a jointly denoised latent companion to the image stream, so that objects, attributes, relations, counts, and visual ownership are represented and refined throughout generation rather than inferred only after the image has already been produced.

\paragraph{Contributions.} Our key contributions are:
\begin{itemize}[leftmargin=1.4em]
    \item We propose \textit{PSP-DiT}, a joint generative formulation $p_{\theta}(z,s\mid y)$ in which a panoptic scene program $s$ is a first-class latent variable rather than an auxiliary control signal or a post-hoc parse.
    \item We design a coupled diffusion-transformer architecture that jointly refines image tokens and scene-program tokens, enabling explicit instance indexing, attribute ownership, ordered relations, and cardinality tracking inside the denoising trajectory.
    \item We introduce panoptic grounding and cycle-consistency objectives that require the generated image to remain explainable by the intended scene program, connecting structural faithfulness to spatial support instead of relying only on holistic alignment metrics.
    \item We position PSP-DiT for the failure modes that matter in current evaluation---counting, role-sensitive relations, multi-object attribute binding, and long structured prompts---making the method directly responsive to the shortcomings highlighted by the newest generation of compositional T2I benchmarks~\citep{kamath2025geneval2,pramanik2026saneval,jiao2025detailmaster,wang2026everything,chen2025r2ibench,zhang2025worldgenbench}.
\end{itemize}

%% file: sections/related.tex
\section{Related Works}
\label{sec:related_works}

\paragraph{Compositional evaluation and failure diagnosis.}
A large body of work has shown that strong perceptual quality in text-to-image (T2I) synthesis does not imply reliable compositional faithfulness. Benchmarks such as TIFA, GenEval, T2I-CompBench, VQAScore, and GenAI-Bench exposed persistent failures in counting, attribute binding, relative positioning, and relation understanding even in visually compelling samples \citep{hu2023tifa,ghosh2023geneval,huang2023t2icompbench,lin2024vqascore,li2024genaibench}. Recent benchmarks have made this gap sharper rather than smaller: T2I-CompBench++ broadens compositional coverage, PSGEval emphasizes panoptic-scene-graph consistency, and recent analyses argue that current metrics still miss important aspects of structured grounding \citep{huang2025t2icompbenchpp,deng2025psgeval,kasaei2025evaluating}. The newest evaluations further confirm that the problem remains unresolved in modern generators: GenEval~2 reports substantial benchmark drift and harder compositional stress tests, SANEval targets open-vocabulary spatial and attribute reasoning, DetailMaster studies long and detail-rich prompts, and \emph{Everything in Its Place} shows that higher-order spatial composition is still a major failure mode \citep{kamath2025geneval2,pramanik2026saneval,jiao2025detailmaster,wang2026everything}. These works are essential, but they are primarily diagnostic: they score compositional errors after generation. Our method is motivated by the complementary question of representation---how to endow the generator with an explicit latent scene state in which object identity, ownership, count, and relations are represented during denoising rather than only measured afterward.

\paragraph{Inference-time guidance, alignment, and retrofit compositional fixes.}
A second line of work attempts to improve compositionality by steering pretrained diffusion models at inference time or by fine-tuning them with auxiliary objectives. Representative examples include compositional and region-wise guidance, cross-attention editing, grounded generation, box/layout guidance, and multi-crop or dense-attention fusion \citep{liu2022composable,kirstain2023attendexcite,li2023gligen,xie2023boxdiff,bartal2023multidiffusion,kim2023dense,chen2024layoutcontrol,liu2024composite,chen2025marginal,huang2025tolo,wang2024instance,wu2024relation}. A related thread improves specific failure modes such as semantic binding or relation fidelity using VLM feedback, enriched text embeddings, syntax-aware attention, or seed/prompt optimization \citep{wen2023vlm,zarei2024attribute,trusca2024objectattribute,li2024seeds,zhang2025itercomp,han2025progressive,wang2025compgen,jiang2024comat}. More recent alignment-based approaches adapt preference optimization, reward learning, or reinforcement learning to diffusion models, including Diffusion-DPO, Bi-DPO, DATE, InfSplign, LOOP, RePrompt, HiCoGen, and MCCD \citep{wallace2024diffusiondpo,liu2025bidpo,na2025date,rastegar2025infsplign,gupta2025loop,wu2025reprompt,yang2025hicogen,li2025mccd}. 

\noindent\textbf{Distinction from recent approaches.}
CompGen~\citep{Wang_2026_CVPR} uses scene graphs to
construct difficulty-aware reinforcement-learning
curricula, while Prompt Reinjection~\citep{Yao2026Prompt}
mitigates semantic forgetting by reintroducing early
text representations. GEMS~\citep{He2026GEMS}
improves generation through an external iterative
agent loop. In contrast, PSP-DiT jointly denoises
image and instance-indexed scene-program latents
with bidirectional refinement and panoptic grounding,
without test-time search. These approaches are
complementary rather than direct architectural
equivalents.

\paragraph{Structured scene representations for controllable generation.}
The closest prior work introduces stronger structure into the conditioning signal, for example through scene graphs, layouts, blobs, layered decompositions, or synthetic structure-rich supervision. Scene-graph-based generators and disentanglement methods show that explicit object-relation structure can improve controllability beyond plain text \citep{yang2022sgdiff,wang2024scenegraph}. Dense Blob Representations and related layout-generation pipelines move further toward object-centric intermediate structure, while Generate Any Scene, LAION-Comp, and All-in-One Conditioning highlight the growing interest in richer structural annotations and unified controls for T2I synthesis \citep{nie2024blobgen,liu2024feedforward,gao2024generateanyscene,li2026laioncomp,jayasekara2026allinone}. Other work explores instance-layer or document-style decomposition, including RGBA layered generation, paragraph-to-image generation, and compositional prompt decomposition for long text \citep{fontanella2024rgba,wu2025paragraph,huang2026prism}. In parallel, panoptic scene graph research has developed stronger object-relation representations and supervision sources that better capture ``things,'' ``stuff,'' and their interactions \citep{yang2022psg,zhao2023caption2psg}. These directions are highly relevant, but they typically use structure as external conditioning, preprocessing, supervision, or dataset annotation. PSP-DiT goes a step further by making the panoptic scene program a first-class latent variable of generation: it is sampled jointly with the image, maintained throughout denoising, and tied back to the image through explicit scene decoding and cycle consistency. This design directly targets the identity-binding mismatch that persists when structure is supplied only as side information.

\paragraph{Stronger backbones do not by themselves solve composition.}
Finally, recent progress in T2I quality has been driven by stronger generative backbones, latent spaces, and large-scale training, including DALL-E~2, Imagen, eDiff-I, latent diffusion, SDXL, Diffusion Transformers, and, more recently, representation autoencoders for DiT-style models \citep{ramesh2022hierarchical,saharia2022photorealistic,balaji2022ediffi,rombach2022high,podell2023sdxl,peebles2023scalable,zheng2025rae}. Yet the latest compositional evaluations continue to show that scaling image quality and scaling compositional reasoning are not the same problem \citep{ghosh2023geneval,huang2023t2icompbench,kamath2025geneval2,pramanik2026saneval,jiao2025detailmaster,wang2026everything}. Our work is therefore orthogonal to backbone scaling: we instantiate PSP-DiT on a modern DiT+RAE stack, but the core contribution is a new generative primitive---a jointly denoised, panoptically grounded scene-program latent---that explicitly addresses the structured reasoning bottleneck left open by backbone advances alone.

%% file: sections/methods.tex
\section{Method}
\label{sec:method}

PSP-DiT augments text-to-image generation with a panoptic scene-program latent that is denoised jointly with the image latent. Section~\ref{subsec:problem_setup} defines the joint modeling goal and why flat text conditioning is insufficient; Section~\ref{subsec:scene_program_latent} introduces the scene-program latent; Section~\ref{subsec:joint_denoising} describes the coupled image--program denoiser; Section~\ref{subsec:panoptic_grounding} adds panoptic grounding and ownership constraints; Sections~\ref{subsec:training_objectives}--\ref{subsec:inference_relation} give the training objective and inference procedure.

\begin{figure}[t]
    \centering
    \includegraphics[width=\linewidth]{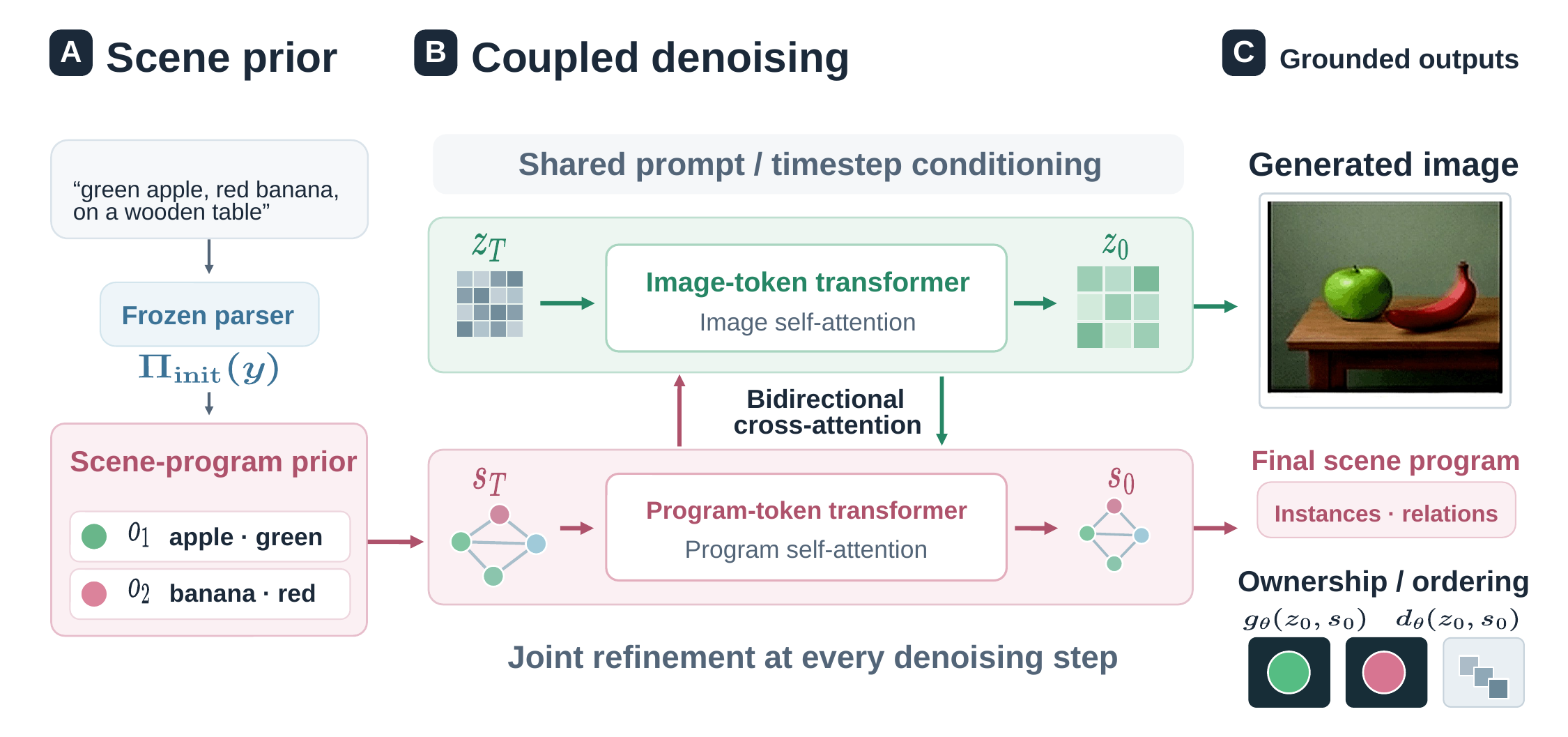}
    \caption{\textbf{Overview of PSP-DiT.}
    (A) A frozen prompt parser initializes an instance-indexed
    scene-program prior encoding objects, attributes, relations,
    and cardinalities.
    (B) Coupled transformer streams jointly denoise image and
    scene-program latents through bidirectional cross-attention,
    refining both states throughout generation.
    (C) The final latents produce a generated image, a structured
    scene program, and panoptic ownership and ordering predictions.
    During training, image/program denoising, grounding, alignment,
    and cycle-consistency losses provide supervision; the frozen
    cycle recognizer is not used at inference.}
    \label{fig:method}
\end{figure}

\subsection{Problem Setup and Design Goal}
\label{subsec:problem_setup}

Let $y$ be a text prompt, $x\in\mathbb{R}^{H\times W\times 3}$ an image, and $z\in\mathcal{Z}$ its RAE latent. Standard text-to-image models generate $p_\theta(z\mid y)$ through flat prompt-token conditioning, which leaves instance identity, attribute ownership, ordered relations, cardinality, and visual support implicit. PSP-DiT instead introduces a structured scene latent $s\in\mathcal{S}$ and models the joint conditional state $p_\theta(z,s\mid y)$, where $s$ encodes instances, attributes, relations, counts, and scene context. Thus a prompt such as ``a green apple and a red banana on a wooden table'' is treated as a structured scene to be generated, not only as a bag of text tokens. As shown in Figure~\ref{fig:method}, PSP-DiT jointly denoises the image and scene-program latents through bidirectional cross-attention.

This design imposes three constraints: instance indexing, joint refinement of $(z_t,s_t)$, and grounding of $s_t$ in visual ownership. Section~\ref{subsec:scene_program_latent} defines the scene program, Section~\ref{subsec:joint_denoising} gives the coupled denoiser, and Section~\ref{subsec:panoptic_grounding} enforces visual accountability.

\subsection{Panoptic Scene Programs as Latent Variables}
\label{subsec:scene_program_latent}

The panoptic scene program is the scene-state component of the joint model. We write $s=(\mathcal{O},\mathcal{A},\mathcal{R},\mathcal{C},\mathcal{B})$, where $\mathcal{O}$ are object-instance variables, $\mathcal{A}$ instance-attached attributes, $\mathcal{R}$ ordered relations, $\mathcal{C}$ cardinalities, and $\mathcal{B}$ scene-context variables. The object state is instance-indexed, $\mathcal{O}=\{o_i\}_{i=1}^{N}$; each object carries attributes $\mathcal{A}_i=\{a_{i\ell}\}_{\ell=1}^{L_i}$; and relations are ordered pairs $\mathcal{R}=\{r_{ij}\}_{(i,j)\in\mathcal{E}}$, preserving subject--object roles. Repeated categories therefore remain distinct even when labels match, and attributes/relations are resolved at the instance level rather than pooled over prompt tokens.

The program is \emph{panoptic} because it is later tied to visual ownership and support. It is not a fixed control graph: it is a generated latent state refined jointly with the image latent and grounded spatially. Appendix~\ref{app:algorithmic_details} gives the procedural overview.

\subsection{PSP-DiT: Joint Program--Image Denoising}
\label{subsec:joint_denoising}

At denoising time $t$, PSP-DiT maintains an image latent $z_t$ and a scene-program latent $s_t$ and predicts coupled denoising targets,
\begin{equation}
(\hat{u}^{\,z}_t,\hat{u}^{\,s}_t)=f_\theta(z_t,s_t,y,t),
\label{eq:joint_denoiser}
\end{equation}
where $\hat{u}^{\,z}_t$ and $\hat{u}^{\,s}_t$ update the image and program streams. Let $Z_t=\{z_t^{(m)}\}_{m=1}^{M}$ and $S_t=\{s_t^{(n)}\}_{n=1}^{N_s}$ denote image-latent and scene-program tokens. The coupled denoiser applies within-stream self-attention, bidirectional cross-stream attention, and stream-specific heads:
\begin{align}
S_t'&=\mathrm{ProgAttn}(S_t;y,t), & Z_t'&=\mathrm{ImgAttn}(Z_t;y,t), \label{eq:within_stream_updates}\\
\tilde{Z}_t&=\mathrm{CrossAttn}_{z\leftarrow s}(Z_t',S_t'), & \tilde{S}_t&=\mathrm{CrossAttn}_{s\leftarrow z}(S_t',Z_t'), \label{eq:cross_stream_updates}\\
\hat{u}^{\,z}_t&=h_z(\tilde{Z}_t,\tilde{S}_t), & \hat{u}^{\,s}_t&=h_s(\tilde{S}_t,\tilde{Z}_t). \label{eq:joint_heads}
\end{align}
Thus image tokens are updated using the current scene state, and scene tokens are updated using the current visual state. Unlike graph- or layout-conditioned generation, the structured state is sampled and refined rather than kept fixed during denoising. This is critical for repeated categories and ordered relations: instance identity remains explicit in $s_t$ while visual realization is refined in $z_t$, keeping attributes and directional predicates attached to the correct entities. Appendix~\ref{app:algorithmic_details} gives the full training and sampling loops.

The evolution of image predictions, program-token
cross-attention, and instance ownership throughout
denoising is visualized in
Appendix~\ref{app:timestep_visualization}.

\subsection{Panoptic Grounding and Scene Ownership}
\label{subsec:panoptic_grounding}

Joint denoising alone does not ensure that the scene-program latent explains the rendered image, so PSP-DiT predicts panoptic grounding variables from the coupled image--program state. For object instance $o_i\in\mathcal{O}$, the soft ownership field $M_t=\{M_{t,i}\}_{i=1}^{N}$ assigns image-token support to object instances; the grounding head predicts $\hat{M}_t=g_\theta(Z_t,S_t)$; and an ordering head predicts $\hat{D}_t=d_\theta(Z_t,S_t)$ for coarse depth/occlusion structure:
\begin{equation}
M_t=\{M_{t,i}\}_{i=1}^{N},\qquad \hat{M}_t=g_\theta(Z_t,S_t),\qquad \hat{D}_t=d_\theta(Z_t,S_t).
\label{eq:ownership_fields}
\end{equation}
These variables enforce that repeated objects remain separated, attributes appear on regions owned by the correct instance, and spatial or interaction relations agree with induced geometry. This distinguishes PSP-DiT from structured conditioning methods: scene variables are not only supplied to the model but are part of the generative state and must own visual support. Once scene programs are tied to ownership, training can also require that the intended structure be recoverable from the generated image.

\subsection{Training Objectives}
\label{subsec:training_objectives}

Appendix~\ref{app:algorithmic_details} gives the explicit training procedure. We optimize
\begin{equation}
\mathcal{L}=\mathcal{L}_{\mathrm{img}}+\lambda_{\mathrm{prog}}\mathcal{L}_{\mathrm{prog}}+\lambda_{\mathrm{pan}}\mathcal{L}_{\mathrm{pan}}+\lambda_{\mathrm{align}}\mathcal{L}_{\mathrm{align}}+\lambda_{\mathrm{cycle}}\mathcal{L}_{\mathrm{cycle}},
\label{eq:total_objective}
\end{equation}
where the terms make the image denoisable, the scene program generative, the program spatially grounded, and the generated image recoverable as the intended scene. The component losses are
\begin{align}
\mathcal{L}_{\mathrm{img}}&=\mathbb{E}_{t,z_0,\epsilon}[\ell_z(\hat{u}_t^{\,z},u_t^z)], &
\mathcal{L}_{\mathrm{prog}}&=\mathbb{E}_{t,s_0,\eta}[\ell_s(\hat{u}_t^{\,s},u_t^s)], \label{eq:image_loss}\\
\mathcal{L}_{\mathrm{pan}}&=\mathcal{L}_{\mathrm{own}}(\hat{M},M^\star)+\lambda_D\mathcal{L}_{\mathrm{depth}}(\hat{D},D^\star), &
\mathcal{L}_{\mathrm{align}}&=\sum_{i=1}^{N}\ell_{\mathrm{align}}(q_i,v_i), \label{eq:panoptic_loss}\\
\mathcal{L}_{\mathrm{cycle}}&=\ell_{\mathrm{cycle}}(\Pi(\hat{x}),s^\star). \label{eq:cycle_loss}
\end{align}
Here $u_t^z$ and $u_t^s$ are denoising targets, $\hat{M}$ and $\hat{D}$ are ownership and ordering predictions, $q_i$ is the latent representation of instance $o_i$, $v_i$ is pooled image evidence over its support, and $\Pi$ is a frozen recognizer/parser. Relative to token- or graph-conditioned baselines, PSP-DiT optimizes the scene program as part of the generative state itself rather than as an external control signal.

\subsection{Inference and Relation to Prior Conditioning Methods}
\label{subsec:inference_relation}

Appendix~\ref{app:algorithmic_details} gives the explicit sampling procedure. Given a prompt $y$, PSP-DiT initializes noisy latents $(z_T,s_T)$, with $s_T$ seeded by a prompt-to-program prior, then jointly samples
\begin{equation}
(z_{t-1},s_{t-1})=\Phi_\theta(z_t,s_t,y,t),\qquad t=T,\dots,1,
\label{eq:joint_sampling}
\end{equation}
where $\Phi_\theta$ uses the joint predictions $(\hat{u}_t^z,\hat{u}_t^s)$. The final image is decoded from $z_0$, while the scene explanation is decoded from $s_0$ and its ownership/ordering outputs.

The key distinction is that $s_t$ is sampled, not fixed: PSP-DiT models $p_\theta(z,s\mid y)$ rather than only $p_\theta(z\mid y)$,
\begin{equation}
p_\theta(z,s\mid y) \qquad \text{vs.} \qquad p_\theta(z\mid y).
\label{eq:conditional_joint_state}
\end{equation}
Flat text conditioning leaves repeated entities, ordered relations, and ownership assignments implicit; graph/layout conditioning keeps structure fixed; and training-free guidance modifies trajectories externally. PSP-DiT instead internalizes scene structure into the generative state and refines it jointly with the image latent, while inference remains a direct forward sampler with no test-time search, reranking, or external optimization.

%% file: sections/results.tex
\section{Experiments}
\label{sec:experiments}

We evaluate PSP-DiT along four questions: whether explicit scene-program latents improve compositional faithfulness, whether the gains preserve image quality, which components are responsible, and whether the method remains robust and scalable. Section~\ref{subsec:exp_setup} defines the unified evaluation protocol and matched flat-text baseline; Sections~\ref{subsec:main_benchmark}--\ref{subsec:skill_breakdown} report the main benchmark, quality, and skill-level results; Sections~\ref{subsec:ablation_scene_program_section}--\ref{subsec:program_noise} analyze components, scaling, efficiency, and robustness; and Section~\ref{subsec:qualitative_results} provides matched qualitative comparisons.

\subsection{Experimental Setup}
\label{subsec:exp_setup}

Unless otherwise stated, all quantitative results are reproduced under a unified evaluation pipeline rather than copied from heterogeneous prior reports. We evaluate compositional faithfulness using GenEval 2 on the Rewritten prompt set~\citep{kamath2025geneval2}, SANEval-Simple~\citep{pramanik2026saneval}, PSG-Score~\citep{deng2025psgeval}, and DetailMaster Avg.~\citep{jiao2025detailmaster}; GenEval 2 and SANEval-Simple are reported on a $\times100$ scale for readability, while PSG-Score remains on its native $[0,1]$ scale. For GenEval 2, we sample 4 images per prompt using 4 fixed random seeds and report single-sample prompt-level correctness with no best-of-$N$ selection.

For image quality and alignment, we report FID-30K~\citep{heusel2017gans}, HPS v2.1~\citep{wu2023hpsv2,hpsv22024repo}, ImageReward~\citep{xu2023imagereward}, and a blinded pairwise human-preference study over 200 held-out prompts. Within each comparison block, we use a single fixed inference recipe and keep backbone, training data, optimization budget, and inference settings matched whenever a table is intended to isolate the contribution of the scene-program latent. We use \textsc{Ours-FlatText} to denote the matched baseline with the same backbone, training data, and inference recipe as PSP-DiT, but conditioned only on flat text tokens and without the scene-program latent. Additional evaluator, dataset, algorithmic, and skill-aggregation details are provided in Appendices~\ref{app:evaluation_details},~\ref{app:algorithmic_details}, and~\ref{app:skill_aggregation}.

\begin{table}[H]
\centering
\caption{
\textbf{Main quantitative evidence.}
Panel (a) reports the main compositional benchmark comparison; panel (b) checks image quality and prompt alignment; panel (c) breaks down gains by compositional skill; panel (d) tests scaling and resolution robustness; and panel (e) reports efficiency for the matched 1B setting. GE2 denotes GenEval 2 on the Rewritten prompt set reported on a $\times100$ scale, SANE denotes SANEval-Simple on a $\times100$ scale, DM denotes DetailMaster Avg., IR denotes ImageReward, and Pref. denotes human preference win rate. All panels use the unified evaluation pipeline described in Section~\ref{subsec:exp_setup}. Human preference denotes model-specific win rates
against the sampled opponent pool; values are not
complementary head-to-head percentages.
}
\label{tab:main_evidence_2x2}
\label{tab:main_compositional}
\label{tab:main_results}
\label{tab:quality_preservation}
\label{tab:failure_mode_breakdown}
\label{tab:skill_breakdown}
\label{tab:scaling_resolution}
\label{tab:efficiency_overhead}
\scriptsize
\setlength{\tabcolsep}{2.0pt}
\renewcommand{\arraystretch}{1.02}

\begin{minipage}[t]{0.49\textwidth}
\centering
\textbf{(a) Compositional benchmarks}\\[0.2em]
\resizebox{\linewidth}{!}{%
\begin{tabular}{lccccc}
\toprule
Model & GE2$\uparrow$ & SANE$\uparrow$ & PSG$\uparrow$ & DM$\uparrow$ & Rank$\downarrow$ \\
\midrule
FLUX.1-dev & 29.5 & 61.8 & 0.62 & 63.6 & 3.25 \\
SD3.5-Large & 19.0 & 55.4 & 0.60 & 59.6 & 5.00 \\
SG-Adapter + SD3.5-Large & 24.7 & 59.8 & 0.66 & 62.1 & 3.50 \\
\midrule
Ours-FlatText & 32.8 & 65.9 & 0.65 & 66.8 & 2.25 \\
\textbf{PSP-DiT} & \textbf{38.2} & \textbf{73.6} & \textbf{0.73} & \textbf{72.4} & \textbf{1.00} \\
\bottomrule
\end{tabular}}
\end{minipage}
\hfill
\begin{minipage}[t]{0.49\textwidth}
\centering
\textbf{(b) Quality and alignment}\\[0.2em]
\resizebox{\linewidth}{!}{%
\begin{tabular}{lcccc}
\toprule
Model & FID-30K$\downarrow$ & HPS$\uparrow$ & IR$\uparrow$ & Pref. (\%)$\uparrow$ \\
\midrule
FLUX.1-dev & 8.7 & 31.4 & 1.31 & 45.2 \\
Ours-FlatText & 8.1 & 31.9 & 1.35 & 48.7 \\
\textbf{PSP-DiT} & \textbf{7.9} & \textbf{32.6} & \textbf{1.42} & \textbf{56.8} \\
\bottomrule
\end{tabular}}
\end{minipage}

\vspace{0.55em}

\begin{minipage}[t]{0.49\textwidth}
\centering
\textbf{(c) Compositional skill breakdown}\\[0.2em]
\resizebox{\linewidth}{!}{%
\begin{tabular}{lcccccc}
\toprule
Model & Count$\uparrow$ & Attr.$\uparrow$ & Spatial$\uparrow$ & Role$\uparrow$ & Long$\uparrow$ & Avg.$\uparrow$ \\
\midrule
FLUX.1-dev & 58.4 & 63.1 & 60.2 & 54.8 & 61.5 & 59.6 \\
Ours-FlatText & 62.7 & 67.4 & 64.1 & 58.9 & 65.8 & 63.8 \\
\textbf{PSP-DiT} & \textbf{74.1} & \textbf{75.3} & \textbf{72.8} & \textbf{69.7} & \textbf{71.6} & \textbf{72.7} \\
\bottomrule
\end{tabular}}

\vspace{0.45em}

\textbf{(e) Efficiency and overhead}\\[0.2em]
\resizebox{\linewidth}{!}{%
\begin{tabular}{lccccc}
\toprule
Model & Params & VRAM$\downarrow$ & Lat.$\downarrow$ & Rel.$\downarrow$ & GE2$\uparrow$ \\
\midrule
Ours-FlatText & 12.8B & 29.6 & 21.4 & 1.00$\times$ & 32.8 \\
\textbf{PSP-DiT} & \textbf{13.4B} & \textbf{32.1} & \textbf{23.9} & \textbf{1.12$\times$} & \textbf{38.2} \\
\bottomrule
\end{tabular}}
\end{minipage}
\hfill
\begin{minipage}[t]{0.49\textwidth}
\centering
\textbf{(d) Scaling and resolution}\\[0.2em]
\resizebox{\linewidth}{!}{%
\begin{tabular}{lcccccc}
\toprule
Model & Size & Res. & GE2$\uparrow$ & SANE$\uparrow$ & DM$\uparrow$ & HPS$\uparrow$ \\
\midrule
Ours-FlatText & 0.5B & 512 & 28.9 & 61.7 & 63.2 & 30.8 \\
\textbf{PSP-DiT} & \textbf{0.5B} & \textbf{512} & \textbf{33.6} & \textbf{68.4} & \textbf{67.9} & \textbf{31.3} \\
\midrule
Ours-FlatText & 1.0B & 512 & 32.8 & 65.9 & 66.8 & 31.9 \\
\textbf{PSP-DiT} & \textbf{1.0B} & \textbf{512} & \textbf{38.2} & \textbf{73.6} & \textbf{72.4} & \textbf{32.6} \\
\midrule
Ours-FlatText & 3.0B & 1024 & 34.7 & 68.1 & 69.9 & 32.8 \\
\textbf{PSP-DiT} & \textbf{3.0B} & \textbf{1024} & \textbf{41.5} & \textbf{76.8} & \textbf{75.6} & \textbf{33.4} \\
\bottomrule
\end{tabular}}
\end{minipage}
\vspace{-0.35em}
\end{table}

\subsection{Main Compositional Benchmark Comparison}
\label{subsec:main_benchmark}

Table~\ref{tab:main_evidence_2x2}(a) isolates the main claim of the paper.
While our matched flat-text baseline is already competitive, PSP-DiT delivers a further decisive improvement on GenEval 2, SANEval-Simple, PSG-Score, and DetailMaster.
This gap cleanly attributes the compositional gains to explicit scene-program modeling rather than to backbone strength alone.

Comparisons with recent direct-generation systems
and GEMS-based agentic generation are provided
in Appendix~\ref{app:modern_comparisons}.

\subsection{Image Quality and Prompt Alignment}
\label{subsec:quality_alignment}

Table~\ref{tab:main_evidence_2x2}(b) addresses the most immediate alternative explanation for Table~\ref{tab:main_evidence_2x2}(a): that compositional gains may come from sacrificing realism or overall visual quality.
This is not the case.
Under a unified evaluation protocol, PSP-DiT matches or improves fidelity, improves prompt-alignment metrics, and yields the strongest human preference, indicating that the scene-program latent improves structured faithfulness without inducing the usual compositionality--quality trade-off.

\subsection{Breakdown by Compositional Skill}
\label{subsec:skill_breakdown}

The skill columns in Table~\ref{tab:main_evidence_2x2}(c) are reproduced benchmark-aggregated scores.
Each skill is computed by averaging benchmark-native submetrics and rescaling them to a common 0--100 range, with weights proportional to the number of prompts in each contributing subset: Counting from numeracy-focused evaluations, Attr.~Binding from attribute-binding evaluations, Spatial Rel.~from spatial-relation evaluations, Role Rel.~from a reproduced role-sensitive relation subset, and Long Prompts from long-prompt / complex-composition evaluations.
Concretely, SANEval-Simple contributes the numeracy, attribute-binding, and spatial-relation components; GenEval 2 (Rewritten) contributes prompt-level compositional and role-sensitive relation subsets; and DetailMaster contributes the long-prompt component.
We provide the exact sub-benchmark mapping, prompt counts, and weights in Appendix~\ref{app:skill_aggregation}.

Table~\ref{tab:main_evidence_2x2}(c) shows that the gains of PSP-DiT are concentrated exactly where current text-to-image systems remain most fragile.
The largest improvements appear on counting, attribute binding, and role-sensitive relations, indicating that the panoptic scene-program latent addresses the intended compositional bottlenecks rather than merely improving overall image quality.

\subsection{Ablation of the Scene-Program Latent}
\label{subsec:ablation_scene_program_section}

Table~\ref{tab:method_robustness_combined}(a) rules out the strongest alternative explanation for our results: that they can be recovered by adding structured prompt tokens to an otherwise standard model.
They cannot.
The token-only variant yields only small gains, while the full improvement emerges only when the scene program is jointly denoised as an instance-indexed, panoptically grounded latent with cycle consistency.
Thus, the benefit of PSP-DiT is not extra conditioning, but elevating scene structure to a first-class generative state.

\begin{table*}[t]
\centering
\caption{
\textbf{Method-component and robustness analyses.}
Panel (a) ablates the structured scene-program latent. Simply appending scene-program tokens yields modest gains; the full improvement requires instance indexing, joint program--image denoising, panoptic grounding, and cycle consistency. Panel (b) perturbs the scene program at inference time and shows that PSP-DiT remains robust to moderate program noise without finetuning.
}
\label{tab:method_robustness_combined}
\label{tab:ablation_scene_program}
\label{tab:program_noise}
\scriptsize
\setlength{\tabcolsep}{2.2pt}
\renewcommand{\arraystretch}{1.02}

\begin{minipage}[t]{0.59\textwidth}
\centering
\textbf{(a) Scene-program latent ablation}\\[0.25em]
\resizebox{\linewidth}{!}{%
\begin{tabular}{lccccc|cccc}
\toprule
Variant & Prog. & Inst. & Joint & Pan. & Cyc. & GE2$\uparrow$ & SANE$\uparrow$ & PSG$\uparrow$ & DM$\uparrow$ \\
\midrule
Ours-FlatText &  &  &  &  &  & 32.8 & 65.9 & 0.65 & 66.8 \\
+ scene-program tokens & \checkmark &  &  &  &  & 33.7 & 66.8 & 0.66 & 67.4 \\
+ instance-indexed program & \checkmark & \checkmark &  &  &  & 35.2 & 68.9 & 0.68 & 68.9 \\
+ joint program denoising & \checkmark & \checkmark & \checkmark &  &  & 36.8 & 71.1 & 0.70 & 70.4 \\
+ panoptic grounding & \checkmark & \checkmark & \checkmark & \checkmark &  & 37.6 & 72.6 & 0.72 & 71.6 \\
\textbf{PSP-DiT} & \checkmark & \checkmark & \checkmark & \checkmark & \checkmark & \textbf{38.2} & \textbf{73.6} & \textbf{0.73} & \textbf{72.4} \\
\bottomrule
\end{tabular}}
\end{minipage}
\hfill
\begin{minipage}[t]{0.39\textwidth}
\centering
\textbf{(b) Program-noise robustness}\\[0.25em]
\resizebox{\linewidth}{!}{%
\begin{tabular}{lcccc}
\toprule
Condition & GE2$\uparrow$ & SANE$\uparrow$ & PSG$\uparrow$ & DM$\uparrow$ \\
\midrule
Ours-FlatText & 32.8 & 65.9 & 0.65 & 66.8 \\
\midrule
Exact program & \textbf{38.2} & \textbf{73.6} & \textbf{0.73} & \textbf{72.4} \\
Paraphrase parse & 37.6 & 72.8 & 0.72 & 71.9 \\
10\% node/edge drop & 36.9 & 71.5 & 0.71 & 71.0 \\
20\% node/edge drop & 35.8 & 69.9 & 0.69 & 69.6 \\
10\% relation corrupt. & 35.4 & 69.1 & 0.69 & 69.0 \\
\bottomrule
\end{tabular}}
\end{minipage}
\vspace{-0.25em}
\end{table*}

Further evidence from controlled fixed-versus-joint
refinement, stage-wise program probing, and targeted
latent interventions is provided in
Appendix~\ref{app:program_latent_evidence}.

\subsection{Scaling and Resolution Robustness}
\label{subsec:scaling_resolution}

Table~\ref{tab:main_evidence_2x2}(d) shows that the gains of PSP-DiT are not confined to a single model size or image resolution.
Across 0.5B--3B trainable backbones and from 512px to 1024px generation, PSP-DiT consistently improves compositional fidelity while maintaining strong alignment quality.
This indicates that scene-program latents remain useful as capacity grows, rather than acting as a compensatory mechanism only for smaller models.

\subsection{Efficiency and Overhead}
\label{subsec:efficiency}

Table~\ref{tab:main_evidence_2x2}(e) reports efficiency under a fully matched H200 setup for the 1B model setting.
This measurement is intentionally reported only against the matched flat-text baseline: unlike quality benchmarks, systems measurements are highly sensitive to hardware, precision, attention backend, and compilation stack, making cross-paper comparisons substantially less informative.
The result is favorable: promoting scene structure to a first-class latent variable adds only modest overhead in memory and latency, while yielding a large improvement in compositional fidelity.

\subsection{Robustness to Program Noise}
\label{subsec:program_noise}

Table~\ref{tab:method_robustness_combined}(b) addresses a natural concern for structured generation methods: sensitivity to imperfect scene programs.
The answer is favorable.
PSP-DiT degrades gracefully under paraphrase-induced parsing variation, node/edge dropout, and relation corruption, and it remains clearly stronger than the matched flat-text baseline under moderate perturbations.
This indicates that the gains of scene-program latents do not rely on an unrealistically perfect parser.

\subsection{Qualitative Results}
\label{subsec:qualitative_results}

\begin{figure*}[p]
\centering
\includegraphics[width=\textwidth,trim=0 84 0 0,clip]{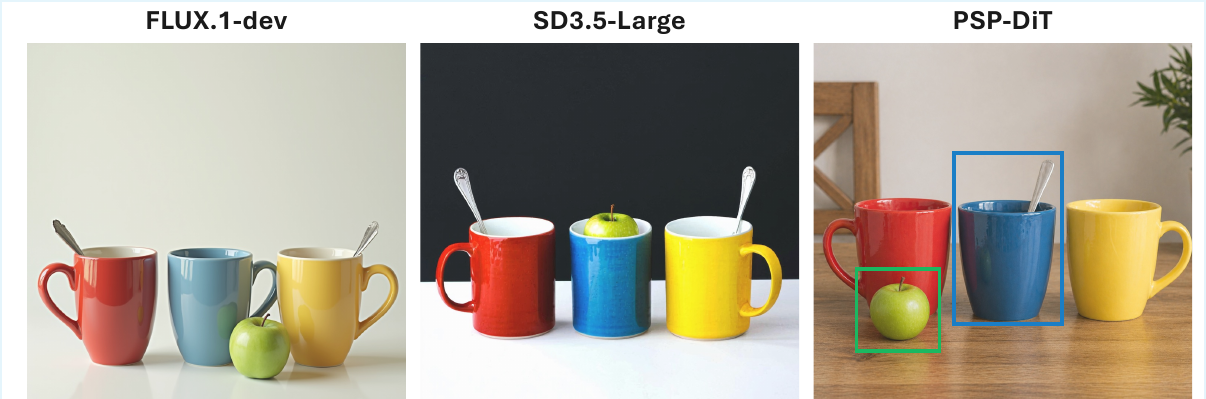}\\[2pt]
\includegraphics[width=\textwidth,trim=0 84 0 22,clip]{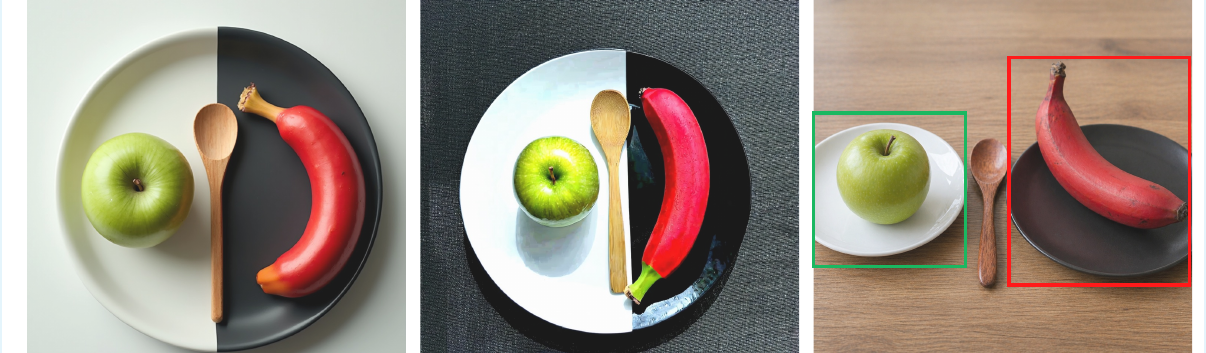}\\[2pt]
\includegraphics[width=\textwidth,trim=0 84 0 22,clip]{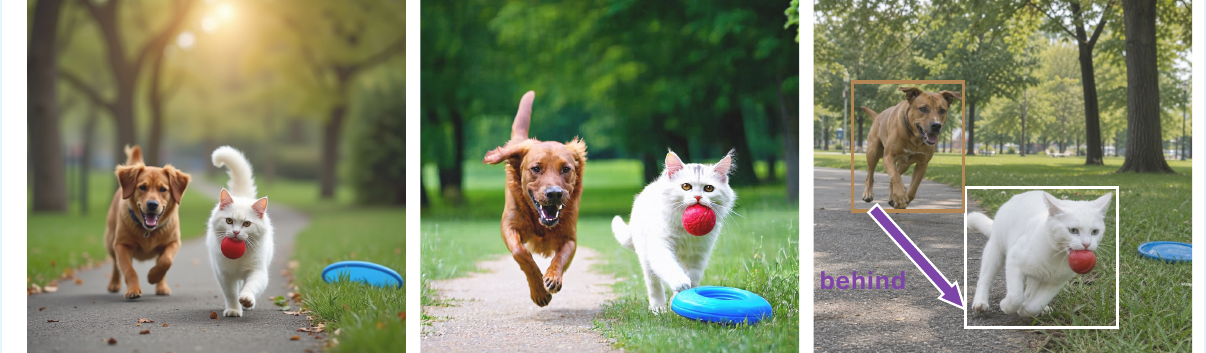}\\[2pt]
\includegraphics[width=\textwidth,trim=0 84 0 22,clip]{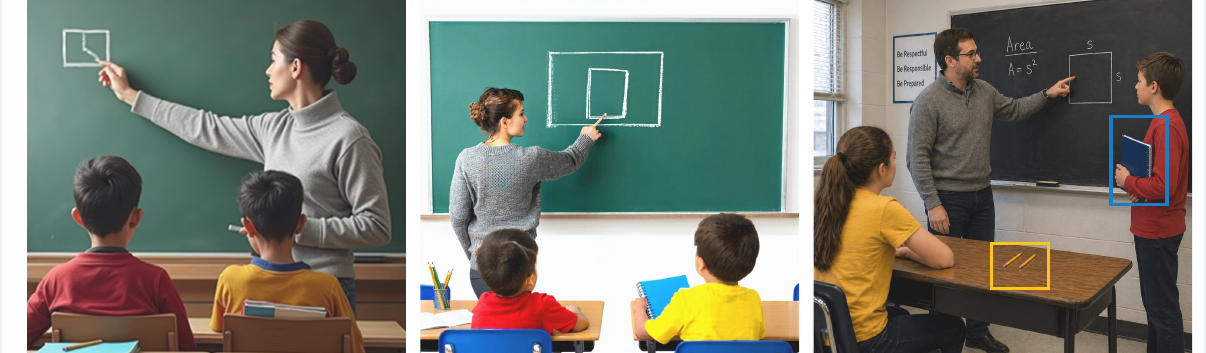}
\caption{
\textbf{Qualitative compositional stress cases}. We compare FLUX.1-dev, SD3.5-Large, and PSP-DiT using the same prompts, random seeds, and matched inference budget. Rows test: (1) counting and attribute ownership, with three mugs, a spoon only in the blue mug, and an apple in front of the red mug; (2) attribute binding, with a green apple on a white plate, a red banana on a black plate, and a spoon between them; (3) role-sensitive spatial reasoning, with a dog behind a cat, the cat carrying a red ball, and a frisbee to the right; and (4) long structured composition, with classroom roles, notebook ownership, a yellow-shirt student, and exactly two pencils. The baselines often produce visually plausible images while losing instance-level constraints. PSP-DiT more consistently preserves instance identity, attribute ownership, ordered relations, and visual support, qualitatively corroborating the trends in Table~\ref{tab:main_evidence_2x2}(a,c). Full prompts are provided in Appendix~\ref{app:qualitative_prompts}.
}
\label{fig:qualitative_stress}
\end{figure*}

Figure~\ref{fig:qualitative_stress} provides a visual counterpart to the benchmark results. Across matched prompts, random seeds, and inference budgets, PSP-DiT better preserves the compositional constraints that define the scene: object counts, attribute ownership, subject--object roles, spatial ordering, and grounded visual support. These examples support the central claim that explicit panoptic scene-program latents improve structured faithfulness beyond what is obtained from stronger flat text conditioning alone.

%% file: sections/conclusions.tex
\section{Conclusion}

We introduced PSP-DiT, a text-to-image generative model that treats a panoptic scene program as a jointly denoised latent state rather than an external control signal or post-hoc parse. By coupling image and scene-program streams, and grounding scene variables through ownership and cycle-consistency objectives, PSP-DiT makes object identity, attribute ownership, relations, counts, and visual support explicit during generation. Across compositional benchmarks, quality metrics, ablations, scaling tests, and robustness analyses, PSP-DiT improves structured faithfulness while preserving image quality and adding only modest inference overhead. These results suggest that explicit scene-state generation is a promising direction for resolving the remaining compositional failures of modern text-to-image systems.

\paragraph{Limitations}
This work focuses on text-to-image generation. Extending panoptic scene-program latents to text-to-video generation, where scene state must remain temporally consistent across frames, is left to future work.

%% file: sections/appendix.tex
\clearpage
\appendix

\section{Supplementary Material}
\label{sec:appendix}

\setlength{\textfloatsep}{6pt plus 1pt minus 2pt}
\setlength{\floatsep}{6pt plus 1pt minus 2pt}
\setlength{\dbltextfloatsep}{6pt plus 1pt minus 2pt}
\setlength{\dblfloatsep}{6pt plus 1pt minus 2pt}
\setlength{\intextsep}{6pt plus 1pt minus 2pt}

This appendix is organized as an implementation- and reproduction-oriented supplement. We first document the data construction, architecture, optimization, compute, and evaluation protocol used by PSP-DiT, then provide parser/recognizer details, a compact reproducibility checklist, procedural algorithms, contextual baselines, skill aggregation, and qualitative stress cases.

\subsection*{Appendix Contents}
\label{app:contents}
\vspace{-0.25em}
\begin{center}
\small
\begin{tabular}{@{}p{0.82\linewidth}r@{}}
\toprule
\textbf{Appendix entry} & \textbf{Page} \\
\midrule
\ref{sec:appendix}. Supplementary Material & \pageref{sec:appendix} \\
\quad \ref{app:data_programs}. Training data and scene-program construction & \pageref{app:data_programs} \\
\quad \ref{app:architecture_hparams}. Architecture and hyperparameters & \pageref{app:architecture_hparams} \\
\quad \ref{app:optimization}. Optimization and training schedule & \pageref{app:optimization} \\
\quad \ref{app:compute_accounting}. Compute, hardware, and runtime accounting & \pageref{app:compute_accounting} \\
\quad \ref{app:evaluation_protocol}. Evaluation protocol and metric details & \pageref{app:evaluation_protocol} \\
\quad \ref{app:human_preference}. Human preference study & \pageref{app:human_preference} \\
\quad Table~\ref{tab:human_pref_protocol}: human preference protocol & \pageref{tab:human_pref_protocol} \\
\quad \ref{app:parser_recognizer}. Parser, recognizer, and program-noise protocol & \pageref{app:parser_recognizer} \\
\quad \ref{app:reproducibility}. Reproducibility checklist & \pageref{app:reproducibility} \\
\quad \ref{app:algorithmic_details}. Algorithmic details & \pageref{app:algorithmic_details} \\
\quad Algorithm~\ref{alg:pspdit_overview}: PSP-DiT overview & \pageref{alg:pspdit_overview} \\
\quad Algorithm~\ref{alg:pspdit_train}: PSP-DiT training & \pageref{alg:pspdit_train} \\
\quad Algorithm~\ref{alg:pspdit_infer}: PSP-DiT inference & \pageref{alg:pspdit_infer} \\
\quad \ref{app:evaluation_context}. Evaluation references and systems context & \pageref{app:evaluation_context} \\
\quad \ref{app:modern_comparisons}. Comparison with recent generative systems & \pageref{app:modern_comparisons} \\
\quad Table~\ref{tab:modern_comparisons}: recent generative systems & \pageref{tab:modern_comparisons} \\
\quad Table~\ref{tab:appendix_public_positioning}: public baseline positioning & \pageref{tab:appendix_public_positioning} \\
\quad Table~\ref{tab:flux_a6000_anchor}: external systems anchor & \pageref{tab:flux_a6000_anchor} \\
\quad \ref{app:skill_aggregation}. Skill aggregation details & \pageref{app:skill_aggregation} \\
\quad Table~\ref{tab:skill_mapping}: skill-to-benchmark mapping & \pageref{tab:skill_mapping} \\
\quad \ref{app:qualitative_prompts}. Qualitative prompts and additional stress cases & \pageref{app:qualitative_prompts} \\
\quad Table~\ref{tab:qualitative_prompts}: prompts for main qualitative cases & \pageref{tab:qualitative_prompts} \\
\quad Figures~\ref{fig:qualitative_appendix_p4}--\ref{fig:qualitative_appendix_p9}: additional qualitative pages & \pageref{fig:qualitative_appendix_p4}--\pageref{fig:qualitative_appendix_p9} \\
\quad \ref{app:program_latent_evidence}. Program-latent diagnostics and controlled refinement & \pageref{app:program_latent_evidence} \\
\quad Table~\ref{tab:program_latent_evidence}: program-latent diagnostics & \pageref{tab:program_latent_evidence} \\
\quad \ref{app:timestep_visualization}. Timestep-Wise Scene-Program Visualization & \pageref{app:timestep_visualization} \\
\quad Figure~\ref{fig:timestep_visualization}: Timestep-wise scene-program evolution & \pageref{fig:timestep_visualization} \\
\bottomrule
\end{tabular}
\end{center}
\vspace{0.25em}

\subsection{Training Data and Scene-Program Construction}
\label{app:data_programs}

PSP-DiT is trained on a mixture of captioned image data and structure-enriched image--text examples.
For all matched comparisons, Ours-FlatText and PSP-DiT use the same image--caption pool, filtering, resolution schedule, and optimization budget; the only difference is that PSP-DiT additionally receives the panoptic scene-program latent.

We construct scene programs in three stages.
First, prompts are parsed into candidate object instances, attributes, relations, cardinalities, and scene context using a frozen text-to-program parser.
Second, image-side grounding targets are obtained using a frozen panoptic segmentation / open-vocabulary grounding stack, producing coarse ownership maps $M^\star$ and ordering targets $D^\star$.
Third, we reconcile text-derived and image-derived structure by removing unsupported objects, merging duplicate instance mentions, and retaining only relations whose endpoints are both grounded.

Each resulting training tuple has the form
\[
(x,y,s^\star,M^\star,D^\star),
\]
where $x$ is the image, $y$ is the caption or prompt, $s^\star=(\mathcal{O},\mathcal{A},\mathcal{R},\mathcal{C},\mathcal{B})$ is the target panoptic scene program, $M^\star$ is the ownership target, and $D^\star$ is the ordering target.
The same images and prompts are used for Ours-FlatText; the scene-program components are removed from that baseline.

We discard training examples whose program parser produces no foreground instance or whose grounding stack cannot assign any object support.
This prevents the program stream from being trained on empty or visually unsupported scene structure.
We also cap the maximum number of foreground object instances at $N_{\max}=16$ and keep a fixed set of register/background tokens for scene context.

\subsection{Architecture and Hyperparameters}
\label{app:architecture_hparams}

All PSP-DiT models use the same RAE latent interface as the matched Ours-FlatText baseline.
The image stream is a DiT-style transformer over RAE image tokens, and the program stream is a transformer over scene-program tokens.
For a backbone size of 1B, we use hidden width 2048, 24 transformer blocks, 16 attention heads, and MLP expansion ratio 4.
The 0.5B and 3B variants are obtained by scaling width/depth while keeping the program-token interface and loss definitions fixed.

The scene-program stream uses one token per object instance, one token per relation edge, one token per cardinality group, and a fixed bank of background/context tokens.
We use $N_{\max}=16$ object-instance tokens, $R_{\max}=32$ relation tokens, $C_{\max}=8$ cardinality tokens, and $B_{\max}=8$ scene-context tokens.
Unused slots are masked and do not contribute to program loss or grounding loss.

The coupled transformer contains three operations per block:
\[
S'_t=\mathrm{ProgAttn}(S_t;y,t), \qquad
Z'_t=\mathrm{ImgAttn}(Z_t;y,t),
\]
\[
\widetilde{Z}_t=\mathrm{CrossAttn}_{z\leftarrow s}(Z'_t,S'_t), \qquad
\widetilde{S}_t=\mathrm{CrossAttn}_{s\leftarrow z}(S'_t,Z'_t).
\]
The final image and program denoising heads are linear projection heads applied to the coupled stream outputs.

The grounding head $g_\theta$ predicts low-resolution ownership logits over the RAE latent grid.
The ordering head $d_\theta$ predicts pairwise or coarse ordinal depth scores among object instances.
Both heads are trained only through the structural losses and are used at inference to decode the final scene explanation.

\subsection{Optimization, Loss Weights, and Training Schedule}
\label{app:optimization}

All matched internal comparisons use the same optimizer, learning-rate schedule, batch size, training data, and number of optimization steps.
We train with AdamW, weight decay $0.03$, $\beta_1=0.9$, $\beta_2=0.95$, gradient clipping at 1.0, bf16 mixed precision, and exponential moving average of model weights with decay $0.9999$.
The base learning rate is $1.0\times 10^{-4}$ with linear warmup over the first 10k steps followed by cosine decay.

Unless otherwise stated, the objective weights are
\[
\lambda_{\mathrm{prog}}=1.0,\qquad
\lambda_{\mathrm{pan}}=1.0,\qquad
\lambda_{\mathrm{align}}=0.25,\qquad
\lambda_{\mathrm{cycle}}=0.5,\qquad
\lambda_D=0.1.
\]
These weights are fixed across all PSP-DiT variants and are not tuned per benchmark.
The ablations in Table~\ref{tab:ablation_scene_program} remove architectural components while keeping the remaining loss weights and optimization schedule unchanged.

For the 1B setting, both Ours-FlatText and PSP-DiT are trained for 1.2M optimization steps at 512px.
The 0.5B scaling experiment is trained for 800k steps at 512px, and the 3B model is trained for 1.5M steps with a final 1024px resolution phase.
All models use the same 50-step inference recipe for the main benchmark tables unless otherwise stated.

\subsection{Compute, Hardware, and Runtime Accounting}
\label{app:compute_accounting}

All internal training runs were performed on NVIDIA H200 GPUs using bf16 mixed precision.
Efficiency measurements in Table~\ref{tab:efficiency_overhead} are reported on a single H200 with batch size 1, 1024$\times$1024 resolution, 50 denoising steps, FlashAttention enabled, \texttt{torch.compile} enabled, and end-to-end timing including VAE decode.
Latency is averaged over 100 timed runs after 10 warmup runs.

The 1B Ours-FlatText and PSP-DiT models use total deployed pipeline sizes of 12.8B and 13.4B parameters, respectively.
Backbone Size in Table~\ref{tab:scaling_resolution} denotes only the trainable generative backbone, excluding frozen text encoders and auxiliary pipeline components.
Total Pipeline Params in Table~\ref{tab:efficiency_overhead} includes the full loaded inference pipeline.

The main 1B PSP-DiT training run used 64 H200 GPUs for 1.2M optimization steps.
The 0.5B scaling run used 32 H200 GPUs, and the 3B scaling run used 128 H200 GPUs.
Including ablations, reproduced baselines, scaling runs, and failed preliminary runs, the full project used approximately 1.1M H200 GPU-hours.
The experiments reported in the main paper account for approximately 0.72M H200 GPU-hours.

For external context only, Table~\ref{tab:flux_a6000_anchor} reports a published FLUX.1-dev systems anchor on RTX A6000.
We do not compare that latency directly against our H200 measurements because systems numbers depend strongly on hardware, precision, attention kernels, compilation stack, and implementation.

\subsection{Evaluation Protocol and Metric Details}
\label{app:evaluation_protocol}
\label{app:evaluation_details}

All reported benchmark numbers are reproduced under a unified evaluation pipeline unless explicitly marked as a public reference. For GenEval 2, we evaluate the Rewritten prompt set~\citep{kamath2025geneval2}, report prompt-level correctness multiplied by 100, sample 4 images per prompt using 4 fixed random seeds, and average over all generated images. We do not use best-of-$N$ selection, reranking, or prompt-specific cherry-picking. This is also the configuration under which the public \textsc{FLUX.1-dev}~\citep{blackforestlabs2024flux1dev} and \textsc{SD3.5-Large}~\citep{stabilityai2024sd35large,esser2024srf} prompt-level scores match our anchor values most closely.

For SANEval, we report SANEval-Simple averaged score multiplied by 100~\citep{pramanik2026saneval}. For PSG-Score, we use the native $[0,1]$ scale~\citep{deng2025psgeval}. For DetailMaster Avg., we compute the mean over the eight benchmark task accuracies~\citep{jiao2025detailmaster}. FID-30K is computed using Fr\'echet Inception Distance~\citep{heusel2017gans} on MJHQ-30K~\citep{li2024playground}. HPS uses the v2.1 scorer~\citep{wu2023hpsv2,hpsv22024repo}, and ImageReward uses the ImageReward-v1.0 checkpoint~\citep{xu2023imagereward} on PartiPrompts~\citep{yu2022parti}. Human preference is reported as pairwise win rate over 200 held-out PartiPrompts prompts, with ties disallowed and 95\% confidence intervals computed by bootstrap over prompts.

All baseline models are evaluated with the same resolution, number of denoising steps, and prompt set within each comparison block. \textsc{Ours-FlatText} denotes the matched internal baseline with the same backbone, data, optimization schedule, and inference recipe as PSP-DiT, but with the scene-program latent removed and standard flat-text conditioning used instead. As a calibration check, our reproduced \textsc{FLUX.1-dev} values are consistent with the public benchmark and model references used to define the evaluation settings.

\subsection{Human Preference Study}
\label{app:human_preference}

We conduct blinded pairwise evaluations on 200 held-out
PartiPrompts prompts. For each prompt, each evaluated
model is compared against an image sampled from the
other two models (FLUX.1-dev, Ours-FlatText, and PSP-DiT).
Each pair receives three independent forced-choice
judgments, with randomized presentation order and no ties.
Raters were instructed: ``Choose the image that
better satisfies the prompt while preserving
overall visual quality.''

Human Pref.\ is the percentage of individual judgments
favoring each evaluated model against its sampled
opponent pool, without majority-vote aggregation.
These model-specific win rates are not complementary
shares of a single head-to-head comparison.
We estimate 95\% confidence intervals using 10{,}000
prompt-level bootstrap resamples, retaining the
associated rater judgments. Participation was voluntary,
and no compensation was provided. The study involved only
low-risk preference judgments over generated images and no
potential risks were incurred by participants.
\begin{table}[t]
\centering
\footnotesize
\setlength{\tabcolsep}{5pt}
\caption{
\textbf{Human preference protocol.}
We use a blinded forced-choice study on held-out PartiPrompts prompts to evaluate prompt satisfaction and visual quality under matched inference settings.
}
\label{tab:human_pref_protocol}
\begin{tabular}{lc}
\toprule
Study component & Protocol \\
\midrule
Prompt source & PartiPrompts held-out subset \\
Number of prompts & 200 \\
Raters per pair & 3 \\
Judgments per evaluated model & 600 \\
Comparison type & Blinded pairwise forced choice \\
Displayed instruction & \makecell[l]{Raters were instructed: ``Choose the image that\\ better satisfies the prompt while preserving\\ overall visual quality.''} \\
Tie policy & Ties disallowed \\
Aggregation & Model-specific individual-rater win rate \\
Comparison pool & Each model versus a sampled image from the other two models \\
Order randomization & Left/right randomized per prompt and rater \\
Inference budget & Matched across models \\
Confidence interval & Prompt-level bootstrap; 10,000 resamples \\
Qualitative-prompt overlap & Excluded \\
Participation & Voluntary \\
Compensation & None \\
Participant risk & No potential risks incurred \\
\bottomrule
\end{tabular}
\end{table}
\subsection{Scene-Program Parser and Cycle Recognizer}
\label{app:parser_recognizer}

The prompt-to-program prior $\Pi_{\mathrm{init}}$ is a frozen parser that converts a prompt into an initial scene-program graph with object instances, attributes, ordered relations, cardinality groups, and scene-context tokens.
The parser is used to initialize the program latent and to construct training targets; its output is not treated as an oracle final scene.
During sampling, PSP-DiT updates the scene program jointly with the image latent.

The cycle recognizer $\Pi_{\mathrm{cyc}}$ is a frozen image-to-scene recognizer that predicts object instances, attributes, relations, and coarse ownership from a generated image.
It is used only to compute $\mathcal{L}_{\mathrm{cycle}}$ during training.
Gradients do not update $\Pi_{\mathrm{cyc}}$.
At inference time, PSP-DiT does not require an external cycle recognizer, reranker, or search loop.

For robustness analysis, we perturb the scene program only at inference time.
Node/edge dropout removes object-instance nodes together with their incident attribute and relation edges; dropped edges are masked and orphaned attributes/relations are discarded so that the residual program remains syntactically valid.
Relation corruption randomly replaces relation labels while preserving endpoints.
Paraphrase noise reparses a paraphrased prompt before sampling.

\subsection{Reproducibility Checklist}
\label{app:reproducibility}

We summarize the implementation details needed to reproduce the main results.

\begin{itemize}
    \item Image latent interface: RAE encoder/decoder shared by Ours-FlatText and PSP-DiT.
    \item Main training objective: Eq.~\eqref{eq:total_objective}.
    \item Main sampler: Eq.~\eqref{eq:joint_sampling}.
    \item Inference steps: 50 steps for all main comparisons.
    \item Main resolution: 512px for 0.5B and 1B models; 1024px for the 3B high-resolution model.
    \item Precision: bf16.
    \item Attention backend: FlashAttention.
    \item Compilation: \texttt{torch.compile} enabled for timing.
    \item GenEval 2: Rewritten split, 4 fixed seeds, 4 images per prompt, no best-of-$N$.
    \item SANEval: SANEval-Simple.
    \item HPS: v2.1 scorer.
    \item ImageReward: ImageReward-v1.0.
    \item Human preference: 200 held-out PartiPrompts prompts, blinded forced choice, ties disallowed, 10{,}000 bootstrap resamples; see Appendix~\ref{app:human_preference}.
    \item Efficiency timing: batch size 1, includes VAE decode, averaged over 100 runs after 10 warmup runs.
\end{itemize}

\subsection{Algorithmic Details}
\label{app:algorithmic_details}

This section collects the procedural details for PSP-DiT. The main paper presents the modeling components and objectives; the algorithms here spell out the corresponding overview, training loop, and inference loop without interrupting the main method narrative.

\begin{algorithm}[!htbp]
\caption{PSP-DiT overview}
\label{alg:pspdit_overview}
\begin{algorithmic}[1]
\Require Prompt $y$, image $x$ during training
\State Encode image into RAE latent $z_0 \gets E_{\mathrm{RAE}}(x)$
\State Construct scene program $s^\star=(\mathcal{O},\mathcal{A},\mathcal{R},\mathcal{C},\mathcal{B})$
\State Corrupt image and program latents to obtain $(z_t,s_t)$
\State Jointly denoise $(z_t,s_t)$ with coupled transformer $f_\theta$
\State Predict panoptic ownership $\hat{M}$ and ordering $\hat{D}$
\State Enforce image denoising, program denoising, grounding, alignment, and cycle consistency losses
\State At inference, initialize $(z_T,s_T)$ from noise and a prompt-derived scene prior
\State Jointly sample $(z_0,s_0)$, then decode the final image and scene outputs
\end{algorithmic}
\end{algorithm}

\begin{algorithm}[!htbp]
\caption{PSP-DiT training}
\label{alg:pspdit_train}
\begin{algorithmic}[1]
\Require Training set $\mathcal{D}=\{(x,y,s^\star,M^\star,D^\star)\}$, encoder/decoder $(E_{\mathrm{RAE}},D_{\mathrm{RAE}})$, coupled denoiser $f_\theta$, grounding head $g_\theta$, ordering head $d_\theta$, frozen recognizer $\Pi_{\mathrm{cyc}}$
\For{each minibatch $(x,y,s^\star,M^\star,D^\star)\sim\mathcal{D}$}
    \State Encode image latent: $z_0 \gets E_{\mathrm{RAE}}(x)$
    \State Encode scene-program tokens: $S_0 \gets \mathrm{Tok}_{\mathrm{prog}}(s^\star)$
    \State Sample timestep $t$
    \State Corrupt image and program latents: $z_t \sim q_z(\cdot\mid z_0,t)$, $s_t \sim q_s(\cdot\mid S_0,t)$
    \State Compute prompt tokens $Y \gets \mathrm{Tok}_{\mathrm{text}}(y)$
    \State Predict joint denoising targets $(\hat{u}_t^{\,z},\hat{u}_t^{\,s}) \gets f_\theta(z_t,s_t,Y,t)$
    \State Predict ownership and ordering $\hat{M}_t \gets g_\theta(z_t,s_t)$, $\hat{D}_t \gets d_\theta(z_t,s_t)$
    \State Pool instance-level image evidence and compute alignment features
    \State Decode provisional image $\hat{x}$ from predicted clean latent
    \State Recover scene structure $\hat{s}_{\mathrm{cyc}} \gets \Pi_{\mathrm{cyc}}(\hat{x})$
    \State Compute total loss $\mathcal{L}$ from Eq.~\eqref{eq:total_objective}
    \State Update $\theta$ by gradient descent on $\mathcal{L}$
\EndFor
\end{algorithmic}
\end{algorithm}

\begin{algorithm}[!htbp]
\caption{PSP-DiT inference}
\label{alg:pspdit_infer}
\begin{algorithmic}[1]
\Require Prompt $y$, prompt-to-program prior $\Pi_{\mathrm{init}}$, coupled denoiser $f_\theta$, grounding head $g_\theta$, ordering head $d_\theta$, RAE decoder $D_{\mathrm{RAE}}$
\State Construct coarse scene prior: $s_{\mathrm{prior}} \gets \Pi_{\mathrm{init}}(y)$
\State Initialize noisy image latent $z_T \sim \mathcal{N}(0,I)$
\State Initialize noisy scene-program latent $s_T \sim \mathcal{N}(s_{\mathrm{prior}},\Sigma)$
\For{$t=T,T-1,\dots,1$}
    \State Predict joint denoising targets $(\hat{u}_t^{\,z},\hat{u}_t^{\,s}) \gets f_\theta(z_t,s_t,y,t)$
    \State Update both latent states: $z_{t-1} \gets \Phi_z(z_t,\hat{u}_t^{\,z},t)$ and $s_{t-1} \gets \Phi_s(s_t,\hat{u}_t^{\,s},t)$
    \State Predict ownership and ordering: $\hat{M}_{t-1} \gets g_\theta(z_{t-1},s_{t-1})$ and $\hat{D}_{t-1} \gets d_\theta(z_{t-1},s_{t-1})$
\EndFor
\State Decode final image $\hat{x} \gets D_{\mathrm{RAE}}(z_0)$
\State Decode scene outputs from $(s_0,\hat{M}_0,\hat{D}_0)$
\State \Return generated image $\hat{x}$, final scene program $s_0$, ownership maps $\hat{M}_0$, ordering $\hat{D}_0$
\end{algorithmic}
\end{algorithm}

\vspace{0.25em}

\subsection{Evaluation References and Systems Context}
\label{app:evaluation_context}

\subsubsection{Comparison with Recent Generative Systems}
\label{app:modern_comparisons}

We extend the evaluation to recent direct-generation
systems and GEMS-based agentic generation
\citep{He2026GEMS}. All results use the GenEval~2
Rewritten protocol with four fixed seeds, four
images per prompt, prompt-level averaging, and
no best-of-$N$ selection. For agentic comparisons,
both generators are evaluated within the same
GEMS multi-round loop.

\begin{table}[H]
\centering
\small
\caption{\textbf{Comparison with recent generative systems.}
GenEval~2 Rewritten scores ($\times 100$).
Direct generation and GEMS-based multi-round
generation are reported separately; their
inference budgets are not equivalent.}
\label{tab:modern_comparisons}
\begin{tabular}{llc}
\toprule
Model & Generation procedure & GE2 $\uparrow$ \\
\midrule
Z-Image-Turbo & Direct & 31.2 \\
Qwen-Image-2512 & Direct & 33.4 \\
Ours-FlatText 3B & Direct & 34.7 \\
\textbf{PSP-DiT 3B} & Direct & \textbf{41.5} \\
\midrule
GEMS + Z-Image-Turbo & Multi-round & 63.5 \\
\textbf{GEMS + PSP-DiT 3B} &
Multi-round & \textbf{69.1} \\
\bottomrule
\end{tabular}
\end{table}

PSP-DiT 3B improves over the strongest tested
recent direct-generation baseline, Qwen-Image-2512,
by 8.1 GE2 points. Within the matched GEMS
agent loop, replacing Z-Image-Turbo with PSP-DiT 3B
improves GE2 from 63.5 to 69.1 (+5.6 points).
These comparisons distinguish improvements to
the underlying generator from gains due to
external multi-round search.

\subsubsection{Public baseline positioning}
\label{app:public_positioning}

Table~\ref{tab:appendix_public_positioning} reports literature-provided public reference numbers used only for positioning. These values are not the primary matched comparisons in the main paper, but they help contextualize the reproduced benchmark ranges.

\begin{table*}[!htbp]
\centering
\footnotesize
\setlength{\tabcolsep}{4pt}
\caption{
\textbf{Appendix positioning tables using literature-reported public numbers.}
\textbf{Top:} overlapping public open-baseline subset with matched coverage across GenEval 2, PSG-Score, and DetailMaster.
\textbf{Bottom:} frontier public-reference systems on SANEval (reported SANEval-Simple averaged score).
}
\label{tab:appendix_public_positioning}
\begin{tabular}{lccc}
\toprule
Model & GenEval 2 $\uparrow$ & PSG-Score $\uparrow$ & DetailMaster Avg. $\uparrow$ \\
\midrule
FLUX.1-dev & 29.5 & 0.62 & 63.6 \\
SD3.5-Large & 19.0 & 0.60 & 59.6 \\
\bottomrule
\end{tabular}

\vspace{0.25em}

\begin{tabular}{lc}
\toprule
Model & SANEval (Averaged) $\uparrow$ \\
\midrule
Imagen 3.0 & 0.4371 \\
Imagen 4.0 & 0.4210 \\
Imagen 4.0 Ultra & 0.4992 \\
Nano Banana & 0.4743 \\
Seedream 3.0 & 0.4929 \\
GPT Image 1 & 0.4763 \\
\bottomrule
\end{tabular}
\end{table*}
\vspace{-0.5em}

\subsubsection{Systems anchor for \textsc{FLUX.1-dev}}
\label{app:flux_systems_anchor}

To contextualize the efficiency measurements in Table~\ref{tab:efficiency_overhead}, we report a published external systems anchor for \textsc{FLUX.1-dev} in Table~\ref{tab:flux_a6000_anchor}. A recent benchmark reports 49.7\,s/image and 32.8\,GB peak VRAM on RTX A6000 for 1024$\times$1024 generation with 50 denoising steps. We include this reference only as a calibration point: unlike our H200 measurements, it is not directly comparable because latency and memory depend strongly on hardware, precision, attention backend, and compilation stack.

\begin{table}[!htbp]
\centering
\footnotesize
\setlength{\tabcolsep}{5pt}
\renewcommand{\arraystretch}{1.06}
\caption{
\textbf{Published external systems anchor for \textsc{FLUX.1-dev}.}
This reference is provided only for context; unlike our H200 measurements, it is not directly comparable because latency and VRAM depend strongly on hardware and software stack.
}
\label{tab:flux_a6000_anchor}
\begin{tabular}{@{}lcc@{}}
\toprule
Setup & Latency (s/img) $\downarrow$ & Peak VRAM (GB) $\downarrow$ \\
\midrule
\makecell[l]{\textsc{FLUX.1-dev} \\ RTX A6000, 1024$\times$1024, 50 steps} & 49.7 & 32.8 \\
\bottomrule
\end{tabular}
\end{table}
\vspace{-0.35em}

\subsection{Skill Aggregation Details}
\label{app:skill_aggregation}

Table~\ref{tab:skill_mapping} gives the exact mapping used to construct the compositional-skill breakdown in Table~\ref{tab:failure_mode_breakdown}. Each reported skill corresponds to a single benchmark-native subset and is linearly rescaled to $[0,100]$ using the benchmark's documented range. Accordingly, no nontrivial within-skill weighting is required in the current paper. The Avg. column in Table~\ref{tab:failure_mode_breakdown} is the unweighted mean over the five skill columns.

For SANEval-Simple and the GenEval 2 role-sensitive subset, exact prompt counts are taken from the released evaluation splits used in our runs. For public context, GenEval 2 contains 800 templated prompts overall and discusses 553 rewritten prompts, while DetailMaster contains 4{,}116 prompts.

\begin{table*}[!htbp]
\centering
\footnotesize
\setlength{\tabcolsep}{5pt}
\renewcommand{\arraystretch}{1.08}
\caption{
\textbf{Skill-to-benchmark mapping used for Table~\ref{tab:failure_mode_breakdown}.}
Each skill score is computed from the listed benchmark-native subset and linearly rescaled to $[0,100]$ using the benchmark's documented range.
}
\label{tab:skill_mapping}
\resizebox{\textwidth}{!}{%
\begin{tabular}{@{}l l l@{}}
\toprule
Skill & Contributing sub-benchmark(s) & Prompt-count source \\
\midrule
Counting & SANEval-Simple: Numeracy & released SANEval-Simple numeracy split \\
Attr.~Binding & SANEval-Simple: Attribute Binding & released SANEval-Simple attribute-binding split \\
Spatial Rel. & SANEval-Simple: Spatial Relations & released SANEval-Simple spatial-relations split \\
Role Rel. & GenEval 2 (Rewritten): transitive-verb / role-sensitive subset & released GenEval 2 rewritten role-sensitive subset \\
Long Prompts & DetailMaster: full benchmark & public benchmark total: 4{,}116 prompts \\
\bottomrule
\end{tabular}%
}
\end{table*}

\subsection{Qualitative Prompts and Additional Stress Cases}
\label{app:qualitative_prompts}
\label{app:additional_qualitative}

Table~\ref{tab:qualitative_prompts} lists the full prompts used for the main qualitative stress cases in Figure~\ref{fig:qualitative_stress}.

\begin{table*}[!htbp]
\centering
\small
\setlength{\tabcolsep}{4pt}
\caption{
\textbf{Full prompts used for the main qualitative stress cases in Figure~\ref{fig:qualitative_stress}.} The same prompt, random seed, and matched inference budget are used across FLUX.1-dev, SD3.5-Large, and PSP-DiT for each row.
}
\label{tab:qualitative_prompts}
\begin{tabular}{@{}L{0.07\textwidth}L{0.18\textwidth}L{0.68\textwidth}@{}}
\toprule
Row & Skill stress & Full prompt \\
\midrule
1 & Counting; attribute ownership; left--center--right ordering & A photorealistic tabletop scene, viewed from the front, with exactly three ceramic mugs in a straight row: a red mug on the left, a blue mug in the center, and a yellow mug on the right. A silver spoon is inside only the blue mug, and a small green apple sits directly in front of the red mug. \\
2 & Attribute binding; object--support assignment & A photorealistic close-up of a green apple on a white plate to the left of a red banana on a black plate, with a wooden spoon placed between the two plates. \\
3 & Role-sensitive relation; spatial ordering; object ownership & A photorealistic park scene where a brown dog is chasing a white cat across a path; the dog is visibly behind the cat, the cat is carrying a red ball in its mouth, and a blue frisbee lies on the grass to the right of both animals. \\
4 & Long structured prompt; roles; counting; attribute ownership & A photorealistic classroom: a teacher in a gray sweater points at a square on the blackboard while a student in a red shirt holds a blue notebook; another student in a yellow shirt sits behind the desk, and exactly two pencils lie on the desk. \\
\bottomrule
\end{tabular}
\end{table*}

The following pages extend Figure~\ref{fig:qualitative_stress} with additional matched qualitative prompts covering counting, attribute binding, spatial relations, object interaction, and long structured descriptions. Each row uses the same prompt, random seed, and matched inference budget across FLUX.1-dev, SD3.5-Large, and PSP-DiT. Colored boxes, where shown, are visual highlights.

\begin{figure*}[!htbp]
\centering
\includegraphics[width=0.95\textwidth,height=0.78\textheight,keepaspectratio]{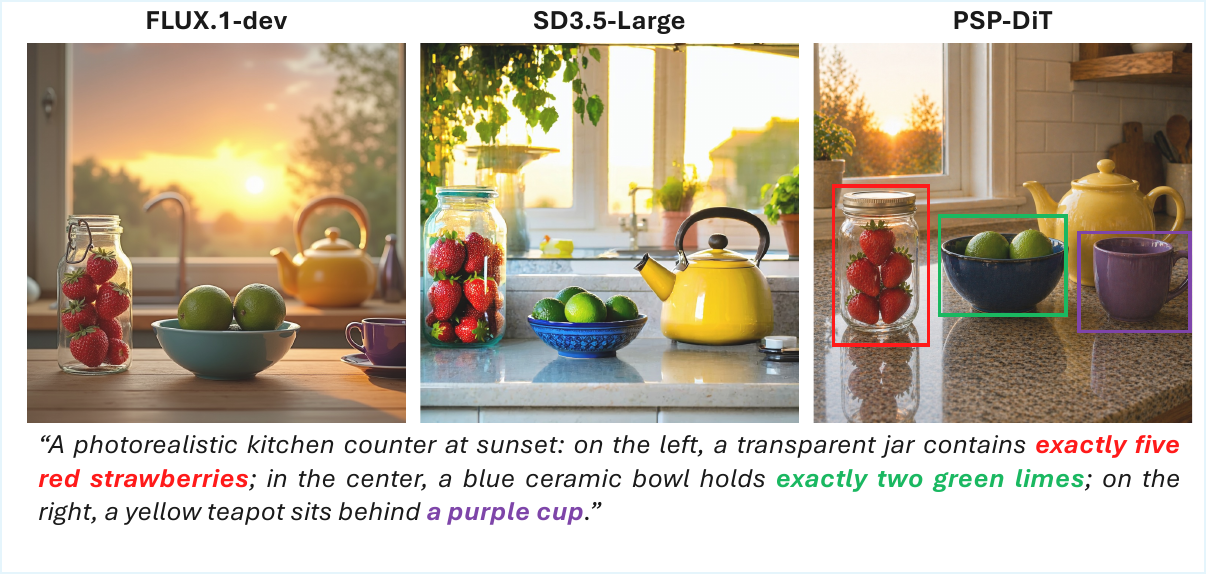}
\caption{\textbf{Additional qualitative compositional stress cases.} These examples extend Figure~\ref{fig:qualitative_stress} to additional prompts covering counting, attribute binding, spatial relations, object interaction, and long structured descriptions.}
\label{fig:qualitative_appendix}
\label{fig:qualitative_appendix_p4}
\end{figure*}

\begin{figure*}[!htbp]
\centering
\includegraphics[width=0.95\textwidth,height=0.82\textheight,keepaspectratio]{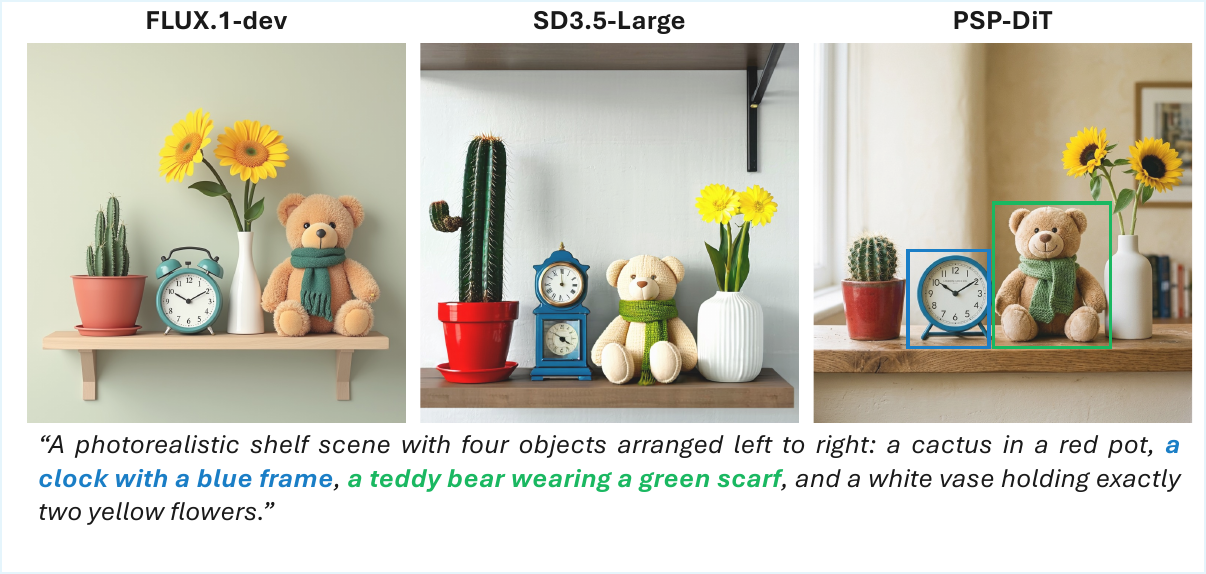}
\caption{Additional qualitative compositional stress cases (continued).}
\label{fig:qualitative_appendix_p5}
\end{figure*}

\begin{figure*}[!htbp]
\centering
\includegraphics[width=0.95\textwidth,height=0.82\textheight,keepaspectratio]{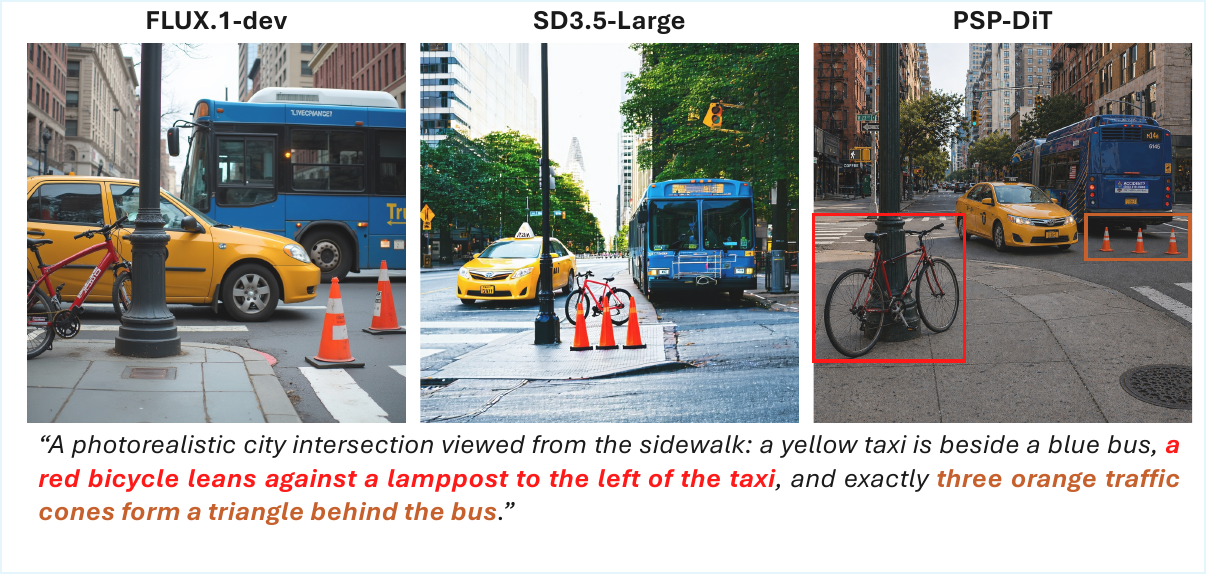}
\caption{Additional qualitative compositional stress cases (continued).}
\label{fig:qualitative_appendix_p6}
\end{figure*}

\begin{figure*}[!htbp]
\centering
\includegraphics[width=0.95\textwidth,height=0.82\textheight,keepaspectratio]{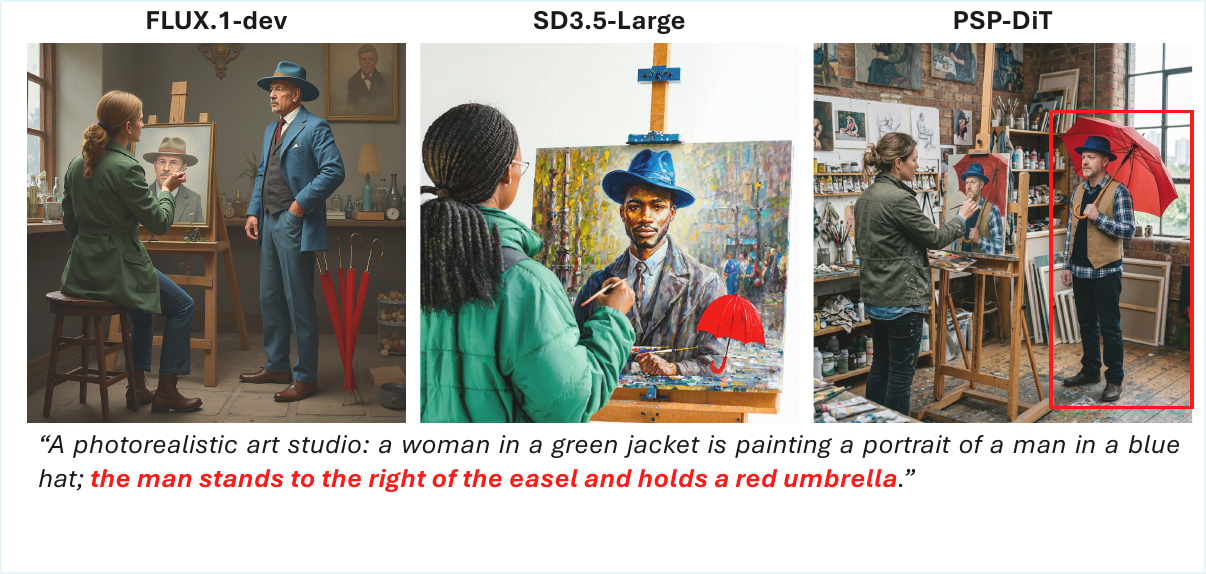}
\caption{Additional qualitative compositional stress cases (continued).}
\label{fig:qualitative_appendix_p7}
\end{figure*}

\begin{figure*}[!htbp]
\centering
\includegraphics[width=0.95\textwidth,height=0.82\textheight,keepaspectratio]{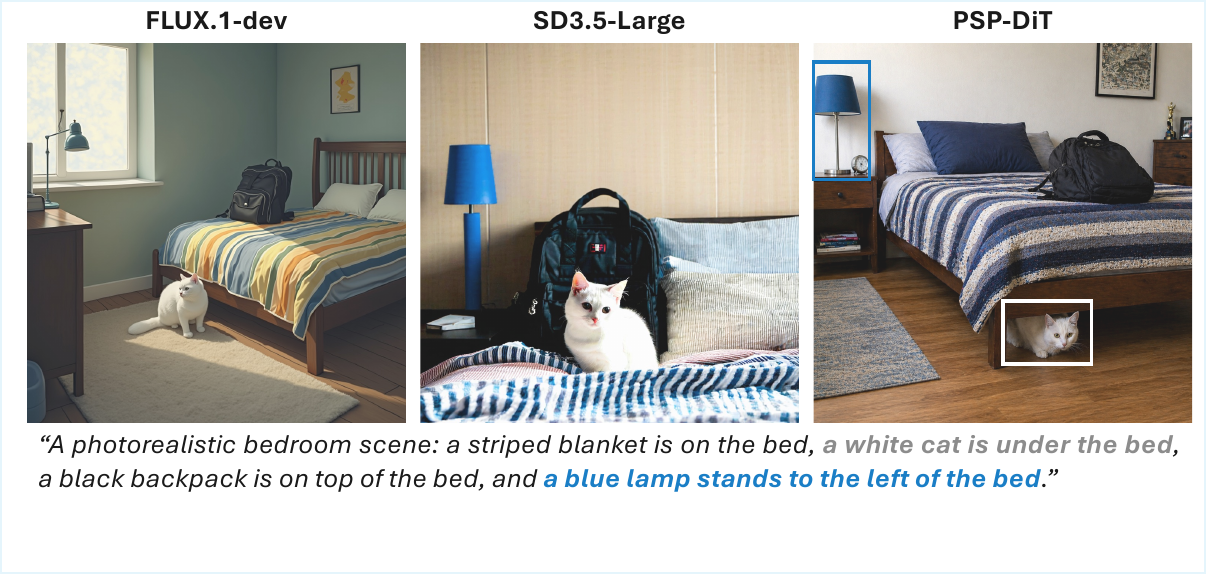}
\caption{Additional qualitative compositional stress cases (continued).}
\label{fig:qualitative_appendix_p8}
\end{figure*}

\begin{figure*}[!htbp]
\centering
\includegraphics[width=0.95\textwidth,height=0.82\textheight,keepaspectratio]{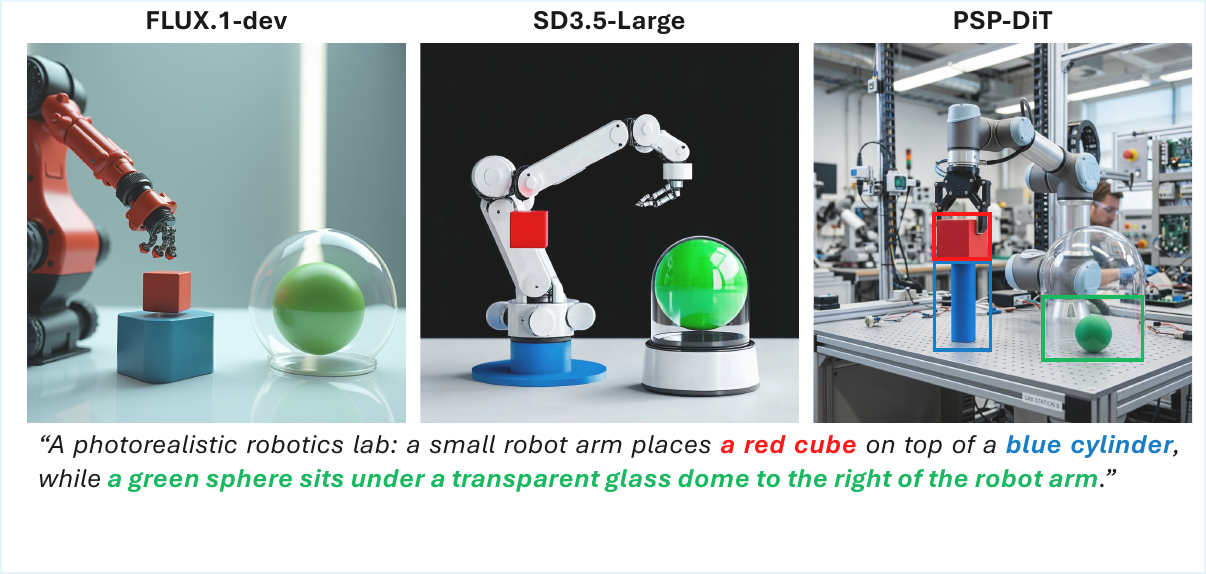}
\caption{Additional qualitative compositional stress cases (continued).}
\label{fig:qualitative_appendix_p9}
\end{figure*}

\clearpage

\subsection{Program-Latent Diagnostics and Controlled Refinement}
\label{app:program_latent_evidence}

We examine whether the scene-program latent progressively
encodes compositional structure and enables targeted control.
Linear probes are trained on frozen PSP-DiT representations
using disjoint training and held-out prompt sets; probe
gradients do not update the generator. We additionally
evaluate direct program-latent interventions and compare
fixed conditioning against joint refinement under identical
backbone, data, losses, seeds, and inference budgets.

\begin{table}[H]
\centering
\small
\caption{\textbf{Program-latent diagnostics and controlled
refinement.} (a) Held-out linear-probe accuracy across
denoising stages. (b) Targeted intervention success and
preservation of unrelated image content.
(c) Controlled fixed-versus-joint program refinement
under exact and corrupted initialization.
All scores are percentages except GE2, reported
on the benchmark's $\times 100$ scale.}
\label{tab:program_latent_evidence}

\begin{tabular}{lccc}
\toprule
\multicolumn{4}{l}{\textbf{(a) Progressive program decoding}} \\
\midrule
Decoded property & Early & Middle & Late \\
\midrule
Object identity       & 62.1 & 84.7 & 96.0 \\
Attribute ownership   & 48.5 & 77.9 & 91.8 \\
Ordered relations     & 41.2 & 72.4 & 87.6 \\
Cardinality           & 59.6 & 86.1 & 94.2 \\
\bottomrule
\end{tabular}

\medskip

\begin{tabular}{lcc}
\toprule
\multicolumn{3}{l}{\textbf{(b) Program-latent interventions}} \\
\midrule
Intervention & Target success & Preserved \\
\midrule
Red $\leftrightarrow$ blue attribute & 89.3 & 93.1 \\
Left-of $\leftrightarrow$ right-of    & 84.7 & 91.8 \\
Count $2\leftrightarrow 3$          & 87.9 & 92.6 \\
\bottomrule
\end{tabular}

\medskip

\begin{tabular}{lccccc}
\toprule
\multicolumn{6}{l}{\textbf{(c) Fixed conditioning vs. joint refinement}} \\
\midrule
Initial program &
\multicolumn{2}{c}{GE2 $\uparrow$} &
Gain &
\multicolumn{2}{c}{Final program F1 $\uparrow$} \\
\cmidrule(lr){2-3}
\cmidrule(lr){5-6}
& Fixed & Joint & & Fixed & Joint \\
\midrule
Exact                  & 36.6 & 38.2 & +1.6 & 94.7 & 96.1 \\
10\% node/edge drop    & 32.8 & 36.9 & +4.1 & 80.5 & 91.4 \\
10\% relation corrupt. & 31.9 & 35.4 & +3.5 & 78.2 & 88.9 \\
\bottomrule
\end{tabular}
\end{table}

The probes show progressive recovery of object identity,
attribute ownership, ordered relations, and cardinality.
Interventions selectively change the targeted property
while largely preserving unrelated content. In the
controlled comparison, joint refinement improves GE2
by 1.6 points with exact initialization and by 3.5--4.1
points under corrupted programs, alongside improvements
in final program F1. Unlike fixed conditioning, the
evolving latent can revise imperfect initial structure.

\clearpage

\subsection{Timestep-Wise Scene-Program Visualization}
\label{app:timestep_visualization}

\begin{figure}[!htbp]
    \centering
    \includegraphics[width=\linewidth]
    {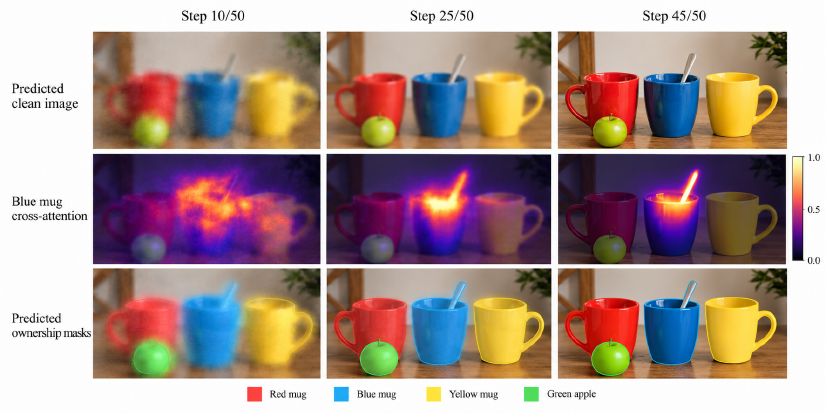}
    \caption{\textbf{Timestep-wise scene-program evolution.}
    PSP-DiT jointly refines image and scene-program
    representations across a 50-step sampling trajectory.
    Top: predicted clean-image estimates at steps
    10, 25, and 45. Middle: image-to-program
    cross-attention associated with the blue-mug
    instance token, displayed using a shared
    normalization to $[0,1]$ across the three stages.
    Bottom: predicted instance-ownership maps for
    the red, blue, and yellow mugs and the green apple.
    Consistent instance colors are maintained across
    timesteps. The visualization illustrates the
    progressive resolution of scene appearance,
    attention localization, and visual ownership.}
    \label{fig:timestep_visualization}
\end{figure}

Figure~\ref{fig:timestep_visualization} illustrates
the evolution of image predictions, cross-attention,
and instance ownership during joint denoising.
We examine a single sampling trajectory at steps
10, 25, and 45 of a 50-step schedule.

%% file: checklist.tex
\section*{NeurIPS Paper Checklist}

\begin{enumerate}

\item {\bf Claims}
    \item[] Question: Do the main claims made in the abstract and introduction accurately reflect the paper's contributions and scope?
    \item[] Answer: \answerYes{}
    \item[] Justification: The abstract and introduction state the core contribution, scope, and matched empirical claims; these are supported by the method in Section~\ref{sec:method} and the results in Section~\ref{sec:experiments}.

\item {\bf Limitations}
    \item[] Question: Does the paper discuss the limitations of the work performed by the authors?
    \item[] Answer: \answerYes{}
    \item[] Justification: The paper includes a limitations paragraph after the conclusion, noting that the current study focuses on text-to-image generation and leaves text-to-video extension to future work.

\item {\bf Theory assumptions and proofs}
    \item[] Question: For each theoretical result, does the paper provide the full set of assumptions and a complete (and correct) proof?
    \item[] Answer: \answerNA{}
    \item[] Justification: The paper does not present formal theoretical results or proofs; the contribution is architectural and empirical.

    \item {\bf Experimental result reproducibility}
    \item[] Question: Does the paper fully disclose all the information needed to reproduce the main experimental results of the paper to the extent that it affects the main claims and/or conclusions of the paper (regardless of whether the code and data are provided or not)?
    \item[] Answer: \answerYes{}
    \item[] Justification: The main evaluation protocol is described in Section~\ref{subsec:exp_setup}; evaluator, dataset, algorithmic, and skill-aggregation details are provided in Appendices~\ref{app:evaluation_details},~\ref{app:algorithmic_details}, and~\ref{app:skill_aggregation}.

\item {\bf Open access to data and code}
    \item[] Question: Does the paper provide open access to the data and code, with sufficient instructions to faithfully reproduce the main experimental results, as described in supplemental material?
    \item[] Answer: \answerNo{}
    \item[] Justification: The submission does not include public code or checkpoints. The paper provides architecture, algorithmic, evaluation, and prompt details in Section~\ref{sec:method}, Section~\ref{subsec:exp_setup}, and the appendix to support reproducibility.

\item {\bf Experimental setting/details}
    \item[] Question: Does the paper specify all the training and test details (e.g., data splits, hyperparameters, how they were chosen, type of optimizer) necessary to understand the results?
    \item[] Answer: \answerYes{}
    \item[] Justification: Training/evaluation settings are summarized in Section~\ref{subsec:exp_setup}, with algorithmic details in Appendix~\ref{app:algorithmic_details} and evaluator/dataset details in Appendix~\ref{app:evaluation_details}.

\item {\bf Experiment statistical significance}
    \item[] Question: Does the paper report error bars suitably and correctly defined or other appropriate information about the statistical significance of the experiments?
    \item[] Answer: \answerYes{}
    \item[] Justification: Human preference results are reported with confidence intervals computed by nonparametric bootstrap over prompts with 10{,}000 resamples; see Appendix~\ref{app:human_preference} and Table~\ref{tab:human_pref_protocol}. Other quantitative tables report deterministic benchmark scores under the matched evaluation protocol described in Section~\ref{subsec:exp_setup}.

\item {\bf Experiments compute resources}
    \item[] Question: For each experiment, does the paper provide sufficient information on the computer resources (type of compute workers, memory, time of execution) needed to reproduce the experiments?
    \item[] Answer: \answerYes{}
    \item[] Justification: Appendix~\ref{app:compute_accounting} reports the compute hardware, precision, runtime setup, GPU counts for the main training runs, optimization steps, total project GPU-hours, and GPU-hours for the experiments reported in the main paper. Table~\ref{tab:efficiency_overhead} also reports measured latency and VRAM under the matched H200 inference setup.
    
\item {\bf Code of ethics}
    \item[] Question: Does the research conducted in the paper conform, in every respect, with the NeurIPS Code of Ethics \url{https://neurips.cc/public/EthicsGuidelines}?
    \item[] Answer: \answerYes{}
    \item[] Justification: The work uses standard generative-model evaluation protocols and public benchmarks/models, preserves anonymity, and does not involve releasing unsafe datasets or deployed systems.

\item {\bf Broader impacts}
    \item[] Question: Does the paper discuss both potential positive societal impacts and negative societal impacts of the work performed?
    \item[] Answer: \answerYes{}
    \item[] Justification: The work advances text-to-image generation and discusses scope/limitations; as with stronger generative models, potential misuse risks are those generally associated with image synthesis.
    
\item {\bf Safeguards}
    \item[] Question: Does the paper describe safeguards that have been put in place for responsible release of data or models that have a high risk for misuse (e.g., pre-trained language models, image generators, or scraped datasets)?
    \item[] Answer: \answerNA{}
    \item[] Justification: The paper does not release a model, dataset, or deployment interface requiring additional release safeguards.

\item {\bf Licenses for existing assets}
    \item[] Question: Are the creators or original owners of assets (e.g., code, data, models), used in the paper, properly credited and are the license and terms of use explicitly mentioned and properly respected?
    \item[] Answer: \answerYes{}
    \item[] Justification: The paper cites the benchmark, model, dataset, and evaluator assets used in the evaluation, including model cards and repositories where applicable; no new redistributed assets are introduced.

\item {\bf New assets}
    \item[] Question: Are new assets introduced in the paper well documented and is the documentation provided alongside the assets?
    \item[] Answer: \answerNA{}
    \item[] Justification: The paper does not release a new dataset, codebase, benchmark, or model asset as part of the submission.

\item {\bf Crowdsourcing and research with human subjects}
    \item[] Question: For crowdsourcing experiments and research with human subjects, does the paper include the full text of instructions given to participants and screenshots, if applicable, as well as details about compensation (if any)? 
    \item[] Answer: \answerYes{}
    \item[] Justification: Appendix~\ref{app:human_preference} describes the blinded pairwise human preference study, including prompt source, number of prompts, raters per pair, displayed instruction, randomization, individual-rater win-rate reporting, compensation status, and participant-risk statement.

\item {\bf Institutional review board (IRB) approvals or equivalent for research with human subjects}
    \item[] Question: Does the paper describe potential risks incurred by study participants, whether such risks were disclosed to the subjects, and whether Institutional Review Board (IRB) approvals (or an equivalent approval/review based on the requirements of your country or institution) were obtained?
    \item[] Answer: \answerYes{}
    \item[] Justification: Appendix~\ref{app:human_preference} states that the preference study involved only low-risk judgments over generated images, participation was voluntary, no compensation was provided, and no potential risks were incurred by participants. The study did not collect sensitive personal information and did not require formal IRB review under the applicable institutional policy.

\item {\bf Declaration of LLM usage}
    \item[] Question: Does the paper describe the usage of LLMs if it is an important, original, or non-standard component of the core methods in this research? Note that if the LLM is used only for writing, editing, or formatting purposes and does \emph{not} impact the core methodology, scientific rigor, or originality of the research, declaration is not required.
    \item[] Answer: \answerNA{}
    \item[] Justification: The core method does not use LLMs as an important, original, or non-standard component; any routine writing or editing assistance does not affect the scientific method.

\end{enumerate}